\documentclass{article}
\usepackage{geometry}
\usepackage{array}
\usepackage[caption=false,font=normalsize,labelfont=sf,textfont=sf]{subfig}
\usepackage{textcomp}
\usepackage{stfloats}
\usepackage{url}
\usepackage{verbatim}
\usepackage{graphicx}
\usepackage{booktabs}
\usepackage{cite}
\usepackage{multirow}
\usepackage{svg}
\usepackage{enumitem}
\usepackage{graphicx} 
\usepackage{times}
\usepackage{soul}
\usepackage{url}
\usepackage{xcolor}
\usepackage[hidelinks]{hyperref}
\usepackage[utf8]{inputenc}
\usepackage{amsmath}
\usepackage{amsthm}
\usepackage[ruled]{algorithm2e}
\usepackage[switch]{lineno}
\usepackage{amsfonts,amssymb,mathrsfs}
\usepackage{rotating}
\usepackage{float}
\usepackage[small,justification=centering]{caption}
\usepackage{orcidlink}
\usepackage[titletoc]{appendix}
\providecommand{\keywords}[1]
{
  \small	
  \textbf{\textit{Keywords--}} #1
}

\title{Deep Reinforcement Learning for Dynamic Algorithm Selection: \\A Proof-of-Principle Study on Differential Evolution}
\date{}

\author{Hongshu Guo~\orcidlink{0000-0001-8063-8984}, Yining Ma~\orcidlink{0000-0002-6639-8547}, Zeyuan Ma~\orcidlink{0000-0001-6216-9379}, Jiacheng Chen~\orcidlink{0000-0002-7539-6156}, Xinglin Zhang~\orcidlink{0000-0003-2592-6945},~\textit{Member,~IEEE}, \\
Zhiguang Cao~\orcidlink{0000-0002-4499-759X}, Jun Zhang~\orcidlink{0000-0001-7835-9871},~\textit{Fellow,~IEEE} and Yue-Jiao Gong~\orcidlink{0000-0002-5648-1160},~\textit{Senior Member,~IEEE}
\thanks{This work was supported in part by the National Natural Science Foundation of China under Grant 62276100, in part by the Guangdong Natural Science Funds for Distinguished Young Scholars under Grant 2022B1515020049, in part by the Guangdong Regional Joint Funds for Basic and Applied Research under Grant 2021B1515120078, and in part by the TCL Young Scholars Program. (Corresponding author: \textit{Yue-Jiao Gong}.)\\
\indent Hongshu Guo, Zeyuan Ma, Jiacheng Chen, Xinglin Zhang, and Yue-Jiao Gong are with the School of Computer Science and Engineering, South China University of Technology, Guangzhou 510006, China (E-mail: 
gongyuejiao@gmail.com)\\
\indent Yining Ma is with the Department of Industrial Systems Engineering and Management, College of Design and Engineering, National University of Singapore, Singapore (E-mail: yiningma@u.nus.edu).\\
\indent Zhiguang Cao is with the School of Computing and Information Systems, Singapore Management University, Singapore. (E-mail: zhiguangcao@outlook.com).\\
\indent Jun Zhang is with Nankai University, Tianjin, China and Hanyang University, Seoul, South Korea and University of Victoria, Melbourne, Australia. (E-mail: junzhang@ieee.org).}
}

\begin{document}



\maketitle

\begin{abstract}
Evolutionary algorithms, such as Differential Evolution, excel in solving real-parameter optimization challenges. However, the effectiveness of a single algorithm varies across different problem instances, necessitating considerable efforts in algorithm selection or configuration. This paper aims to address the limitation by leveraging the complementary strengths of a group of algorithms and dynamically scheduling them throughout the optimization progress for specific problems. We propose a deep reinforcement learning-based dynamic algorithm selection framework to accomplish this task. Our approach models the dynamic algorithm selection a Markov Decision Process, training an agent in a policy gradient manner to select the most suitable algorithm according to the features observed during the optimization process. To empower the agent with the necessary information, our framework incorporates a thoughtful design of landscape and algorithmic features. Meanwhile, we employ a sophisticated deep neural network model to infer the optimal action, ensuring informed algorithm selections. Additionally, an algorithm context restoration mechanism is embedded to facilitate smooth switching among different algorithms. These mechanisms together enable our framework to seamlessly select and switch algorithms in a dynamic online fashion. Notably, the proposed framework is simple and generic, offering potential improvements across a broad spectrum of evolutionary algorithms. As a proof-of-principle study, we apply this framework to a group of Differential Evolution algorithms. The experimental results showcase the remarkable effectiveness of the proposed framework, not only enhancing the overall optimization performance but also demonstrating favorable generalization ability across different problem classes.
\end{abstract}

\keywords{Algorithm selection, deep reinforcement learning, meta-black-box optimization, black-box optimization, differential evolution}

\section{Introduction} \label{Sec:intro}
Evolutionary Computation (EC) has experienced considerable development in the past few decades~\cite{slowik2020evolutionary}. 
Many EC paradigms, such as Genetic Algorithm (GA)~\cite{holland1992adaptation}, Differential Evolution (DE)~\cite{storn1997differential}, Particle Swarm Optimization (PSO)~\cite{kennedy1995particle}, and their variants, have exhibited good performance in solving Black Box Optimization (BBO) problems, owing to their outstanding global exploration and local exploitation abilities. 
However, it has long been observed that for 
different benchmark or practical problems, when several ECs are available, no single one can dominate on all problem instances
, known as 
the ``No-Free-Lunch'' (NFL) theorem~\cite{wolpert1995no}.

To address this issue, an intuitive idea is to leverage the performance complementarity among a group of different EC algorithms, or different configurations for a single EC algorithm, which can serve as different specialties for respective problem instance. This attracted many research efforts to investigate 
which algorithm or algorithm configuration could be promising 
in dealing with 
a particular problem instance~\cite{kerschke2019automated}. We have carefully sorted out the related studies and roughly divided them into three categories: Algorithm Portfolio (AP), Algorithm Configuration (AC) and Algorithm Selection (AS). 

AP, 
also known as \textit{parallel algorithm portfolio}, runs all candidates from a given algorithm set in parallel to determine the most outstanding one for tackling a given problem instance. It was first proposed by Huberman et al.~\cite{huberman1997economics} and then improved by Gomes et al.~\cite{gomes2001algorithm} on the necessary conditions when the parallel portfolio outperforms its component solvers. 
Although AP algorithms benefit from the development of parallel hardware, their huge computation requests are inevitable. 

AC refers to 
finding a proper configuration of an algorithm in order to maximize its performance across a set of problem instances. Classical AC algorithms target an united parameter configuration for all problem instances. For example, 
Ansotegui et al.~\cite{ansotegui2009gender} took advantage of the iterated searching algorithms
to search proper parameter settings in the configuration space. Recently, the AC paradigms dynamically adjusted the algorithmic settings not only for each problem instance but also at each time step during optimization. 
Typically, the reinforcement learning (RL) methods were adopted to learn an online policy for parameter control, such as in the studies of Xue et al.~\cite{xue2022multi} and Biedenkapp et al.~\cite{biedenkapp2020dynamic}. %
Generally, AC provides a flexible way to configure an efficient algorithm to different problem instances. But since 
the configuration space of AC could be extremely large by containing infinite sets of algorithmic components, so far the configuration efficiency remains a challenge~\cite{kerschke2019automated}. 

One way to relieve the above issue of AC is to reduce the configuration space by using expert knowledge to pre-determine a finite set of parameter and operator combinations with guaranteed performance. This is typically known as the AS, which 
targets at selecting the most promising expert-designed algorithm for a given problem. 
Existing AS on real-parameter optimization tasks, such as those proposed by Bischl et al.~\cite{bischl2012algorithm} and 
Kerschke et al.~\cite{kerschke2019automated}, both probed the problem with a few samplings and thereafter calculated informative landscape features. These features then served as input for a machine learning model
that decided which algorithm should be applied to the given problem.

Nonetheless, existing AS works still face the following challenges. 
First, some existing studies focus on static AS that, once an algorithm is selected, it will be executed for the whole optimization process. 
Such methods neglect the utilization of the performance complementary among candidates to achieve comprehensively better performance than the best single algorithm in the candidate set. 
Second, existing AS studies adopt supervised learning methods to estimate candidates' performance based on analytical features and take the estimation results as the basis for algorithm selection.
However, without modeling the optimization progress in detail, the available features for characterizing the problem landscape and/or algorithm behavior are rather limited.

To address the above issues, this paper proposes a deep \textbf{R}einforcement \textbf{L}earning based \textbf{D}ynamic \textbf{A}lgorithm \textbf{S}election (\textbf{RL-DAS}) framework, presenting following key contributions.

\begin{figure}[t]
\centering
\includegraphics[width=0.65\columnwidth]{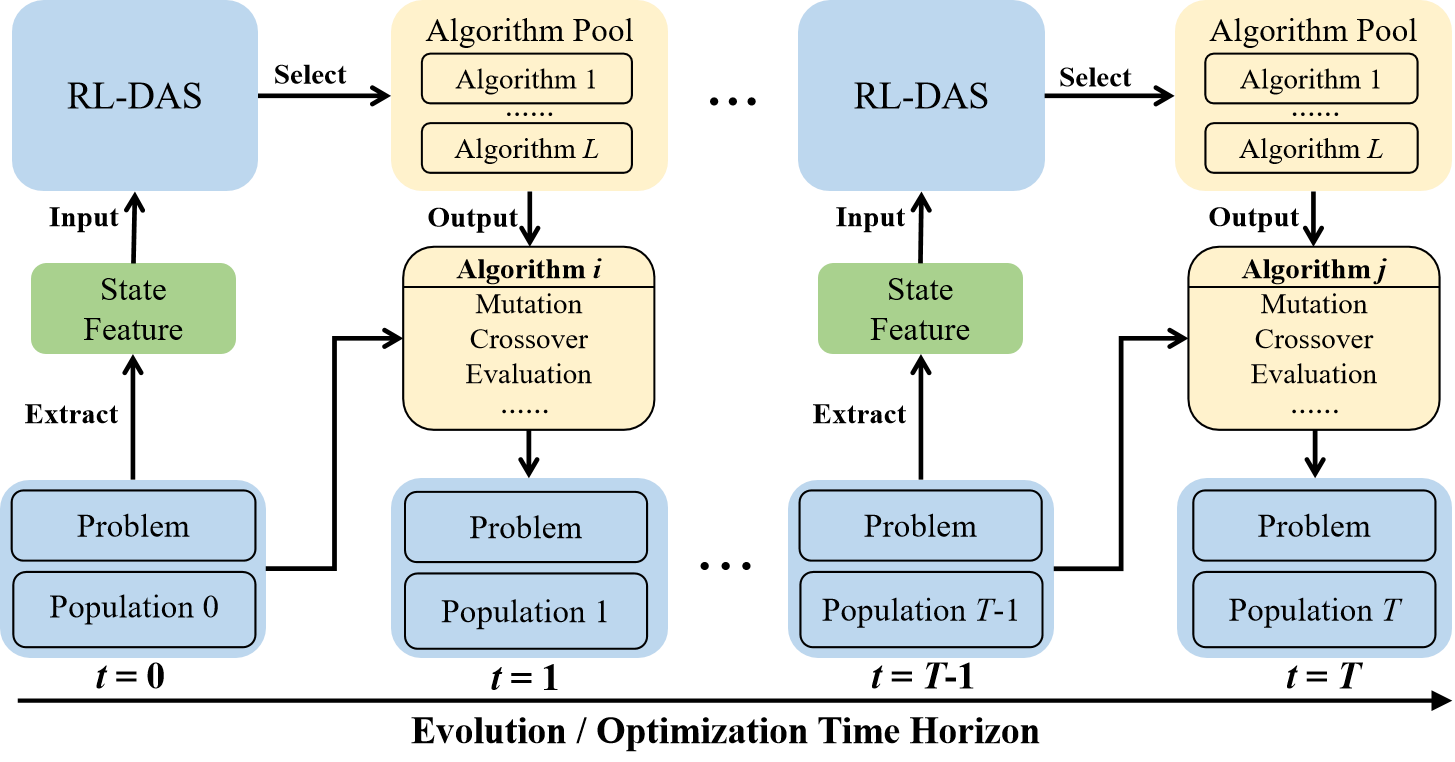}
\caption{The overall structure of RL-DAS.}
\label{dasrl}
\end{figure}

\begin{enumerate}
    \item We extend the AS to a Dynamic Algorithm Selection (DAS) setting that  supports dynamically scheduling different algorithms upon request throughout the optimization process. 
    As illustrated in Fig.~\ref{dasrl}, the optimization process is divided into time intervals. At each interval, an algorithm is dynamically selected to improve the population, taking into account the problem state and the performance of previously selected algorithms. This ensures that the overall optimization outcome is maximized. 
    \item Since our objective is to find the best sequential selections for different problems, we formulate the DAS as a Markov Decision Process (MDP) for sequential decision making. We then leverage the Deep Reinforcement Learning (DRL) method to solve the formulated MDP. 
   Meanwhile, RL-DAS supports the \textit{warm start} 
   when switching between different algorithms by maintaining an algorithm context memory and applying an algorithm restoration step, ensuring a smooth switching process.
    \item Furthermore, in order to fully inform the DRL agent and guide it to make wise decisions, we design a set of simple yet representative features to capture both landscape and algorithmic properties. A deep neural network model is then carefully designed to infer the best action based on the derived features. Our designs allow RL-DAS to generalize to various BBO problems.
    \item We apply RL-DAS on the augmented CEC2021 competition benchmark, 
    with a number of DE algorithms as candidates. The experimental results show that RL-DAS not only outperforms all candidates in the algorithm pool but also exhibits promising zero-shot generalization on unseen problems. 
\end{enumerate}

The rest of this paper is organized as follows: Section~\ref{Sec:pre} presents the preliminaries on DRL and DE. Section~\ref{Sec:method} introduces the detail of RL-DAS
. Section~\ref{Sec:case} is the proof-of-principle study of applying RL-DAS on advanced DE algorithms. The experimental results are depicted in Section~\ref{Sec:Exp} with detailed analysis, followed by a conclusion in Section~\ref{Sec:conclude}.


\section{Preliminaries and Related Works}\label{Sec:pre}

In this section, we briefly introduce the Markov Decision Process and the advanced variants of DE. Table~\ref{tb:abbr} summarizes the abbreviation indices for easy reference.
\begin{table}[t]
\centering
\caption{The abbreviation indices.}
\resizebox{0.5\columnwidth}{!}{%
\begin{tabular}{c|l||c|l}
\hline
Abbr 
& Note
& Abbr
& Note
\\ \hline
EC & Evolutionary Computation
& DAS & Dynamic Algorithm Selection
\\
BBO & Black-Box Optimization
& RL & Reinforcement Learning
\\
DE & Differential Evolution
& DRL & Deep Reinforcement Learning
\\
PSO & Particle Swarm Optimization
& MDP & Markov Decision Process
\\
AP & Algorithm Portfolio
& PPO & Proximal Policy Optimization
\\
AC & Algorithm Configuration
& LA & Landscape Analysis information
\\
AS & Algorithm Selection
& AH & Algorithm History information
\\ \hline
\end{tabular}%
}
\label{tb:abbr}
\end{table}
\subsection{Markov Decision Process and Policy Gradient}

A Markov Decision Process (MDP) could be denoted as $\mathcal{M} := <\mathcal{S}, \mathcal{A}, \mathcal{T}, R>$. Given a state $s_t \in \mathcal{S}$ at time step $t$, 
an action $a_t \in \mathcal{A}$ is selected and interacts with the environment where the next state $s_{t+1}$ is produced through the environment dynamic $\mathcal{T}(s_{t+1} | s_t, a_t)$. A reward function 
$R: \mathcal{S} \times \mathcal{A} \rightarrow \mathbb{R}$ is calculated to measure the performance improvement in the transition. 
Those transitions between states and actions achieve a trajectory $\tau := (s_0, a_0, s_1, \cdots, s_T)$. The target of MDP is to find an optimal policy $\pi^*$ such that the returns (accumulated rewards) in trajectories $\tau$ can be maximized:
\begin{equation}
    \pi^* = \mathop{\arg\max}\limits_{\pi \in \Pi}\sum_{t=0}^T \gamma^{t-1}R(s_t, a_t)
\end{equation}
where $\Pi: \mathcal{S} \rightarrow \mathcal{A}$ selects an action with a given state, $\gamma$ is a discount factor and $T$ is the length of trajectory. 
One method specialized in solving MDPs is the \textit{Policy Gradient}-based DRL, which parameterizes the policy by a deep model $\pi_\theta$ with parameters $\theta$. The policy is expected to decide on a proper action at each state in order to achieve a high return of the trajectory. 
The objective function is formulated as:
\begin{equation}
J(\theta ):=\mathbb{E}_{\tau \sim p_{\theta}(\tau )}[R (\tau )]
\end{equation}
which is maximized by the gradient ascent with gradient:
\begin{equation}\label{pg}
      \frac{\partial J}{\partial \theta} = \mathbb{E}_{\tau\sim p_\theta(\tau)}\left [ \left(\sum_{t=0}^{T}{\nabla_\theta \log\pi_\theta(a_t | s_t)} \right) \left(\sum_{t=0}^{T}{R(s_t, a_t)} \right)\right ]
\end{equation}
In this paper, we adopt a powerful Policy Gradient variant, the Proximal Policy Optimization (PPO)~\cite{schulman2017proximal}, to solve the MDP. It proposes a novel objective with clipped probability ratios, which forms a first-order estimate (i.e., lower bound) of the policy's performance. The algorithm alternates between sampling data from the policy and performing several epochs of optimization on the sampled data to improve data efficiency and ensure reliable performance. Its objective function is:
\begin{equation}\label{ppo_obj}
L_\pi(\theta ):=\mathbb{E}\left[ \min (\eta (\theta )\hat{A},\mathrm{clip(}\eta (\theta ),1-\epsilon ,1+\epsilon )\hat{A}) \right] 
\end{equation}
where $\eta$ is the ratio of the probabilities under the new and old policies which performs the importance sampling, and $\hat{A}$ is the estimated advantage calculated as the difference between the target return $G$ and the estimated return $\hat{G}$. Note that PPO is trained in an actor-critic manner that  a critic network $v_\phi$ is adopted to estimate the expected return $\hat{G}$ for different states. The loss function of $v_\phi$ is formulated as:
\begin{equation}
    L_v(\phi) := \text{MSE}(G, \hat{G})
\end{equation}

\subsection{DE and Its Advanced Variants}\label{Sec:DE}

As mentioned earlier, this paper uses DE as a case study for our RL-DAS framework. Therefore, in this subsection, we review some important DE variants. 
Generally, DE is a high-profile field in EC since proposed by Storn and Price~\cite{storn1997differential}. The DE variants are frequent participants and winners in major BBO competitions~\cite{cec2021TR} and widely applied in various practical applications~\cite{gong2018finding,gong2019real,deng2020enhanced}.
Considering the relevance of DE variants to our study, we categorize them into three distinct groups for review.

\subsubsection{Adaptive DE}
In the literature, there have emerged different DE mutation forms, such as DE/rand/1 and DE/best/1. According to the NFL theorem, no single operator can fit all problem instances, creating room for operator adaptation. 
EPSDE~\cite{mallipeddi2011differential} assigned each individual an operator which would be randomly re-selected if it failed to improve the individual.
LADE~\cite{li2023enhancing} divided the population into leader population and adjoint population in each generation and employed different mutation operators on them for different searching strategies.
Parameters such as the step length $F$ of mutation and the crossover rate $Cr$ also play a significant role in optimization. For instance, jDE~\cite{brest2006self} randomly generated these parameters and remembered those that successfully improved individuals for later generations. 
JDE21~\cite{brest2021self} made further improvements on jDE which split the population into two parts to perform different search strategies. 
SHADE~\cite{tanabe2013success} adopted two memories for $F$ and $Cr$ to record their statistical information instead of remembering their values. Its variants, 
NL-SHADE-RSP~\cite{stanovov2021nl} employed an adaptive mutation operator with an archive of eliminated individuals to increase the population diversity,
NL-SHADE-LBC~\cite{stanovov2022nl} further incorporated a modified bound constraint handling technique and improved the performance.
MadDE~\cite{biswas2021improving} adopted the parameter memories proposed in SHADE and employed three adaptive mutation operators and two crossover operators together with the archive design. 
The JDE21, MadDE, and NL-SHADE-RSP are winners of the CEC2021 competition. 
However, the adaption mechanisms rely heavily on designers' expert knowledge and introduce additional parameters (such as the sizes of memories~\cite{stanovov2022nl} and the probabilities of selecting operators~\cite{biswas2021improving}). These new mechanisms and parameters substantially improve the complexity of algorithm fine-tuning, making the algorithms susceptible to overfitting on specific optimization problems.

\subsubsection{Ensemble DE}
Since different DE variants possess strengths in specific types of optimization problems, several studies directly combined multiple DE variants into an ensemble to enhance overall optimization performance. This approach has resulted in the development of Ensemble DE, which bears high relevance to our study. 
For instance, 
EDEV~\cite{wu2018ensemble} proposed a multi-population based framework integrating JADE~\cite{zhang2009jade}, CoDE, and EPSDE, where the best-performing one on a small indicator subpopulation was assigned a large reward population for a period. In contrast to multi-population methods, HMJCDE~\cite{li2016novel} used a hybrid framework with modified JADE and modified CoDE on a single population, operating them alternately according to the fitness improvement rate. 
While existing Ensemble DE algorithms have consistently outperformed their baselines, showcasing impressive performance, their algorithm selection mechanisms are also manually designed and hence suffer from similar drawbacks of Adaptive DE. Additionally, they often choose candidates based on the historical performance~\cite{wu2018ensemble,li2016novel}. However, for different optimization stages, the best-fitted algorithm differs. An exploratory algorithm might excel in the early optimization phase but could be less advantageous for convergence later on. The selection process could benefit from more nuanced approaches, such as using a machine learning model to learn the relation between optimization states and algorithm selection.

\subsubsection{RL-Assisted DE}

While adaptive and ensemble DE algorithms 
have shown significant progress, the dependence on manual mechanisms and expert knowledge could potentially hinder their further development. Consequently, some researchers have turned to machine learning methods, particularly Reinforcement Learning techniques, in search of breakthroughs~\cite{li2023evolutionary}. Research in this field primarily concentrates on DE configuration, including the selection of mutation and crossover operators, as well as the values of $F$ and $Cr$.
For operator selection, TSRL-DE~\cite{liao2023two}, 
RLDMDE~\cite{yang2023dynamic},
DEDQN~\cite{dedqn} and DEDDQN~\cite{deddqn} employed Q-Learning~\cite{qlearning} and Deep Q-Learning agents~\cite{dqn,ddqn} to select mutation operators for each individual in every generation
 and promote the backbone algorithm performance on several CEC benchmark problems.
Regarding parameter tuning, LDE~\cite{lde} utilized 
LSTM~\cite{graves2012long} to determine the $F$ and $Cr$ values for each individual, based on the states consisting of fitness values and statistics.
Furthermore, some researchers sought more comprehensive control by combining both operator and parameter control, such as the RLHPSDE~\cite{rlhpsde}. 
Current RL-Assisted DE algorithms primarily concentrate on adapting configurations within individual DE algorithms. Given there are plenty of well-performed DE algorithms for different problems, it would be appealing to adopt the RL agents for dynamic algorithm selection, which is, however, a research gap so far. Additionally, the current RL-assisted works arise from the small size of training and testing sets~\cite{dedqn,deddqn,lde}, which could adversely impact their generalization ability.



\section{Methodology}\label{Sec:method}

\subsection{Dynamic Algorithm Selection}

Given a set of algorithms $\Lambda = \{\Lambda_1, \Lambda_2, \cdots, \Lambda_L\}$ and a set of problem instances $I = \{I_1, I_2, \cdots, I_M\}$, the objective of traditional AS is to find an optimal policy as follows,
\begin{equation}
    \pi^* =  \mathop{\arg\max} \limits_{\pi \in \Pi} \sum_{j=1}^M \mathcal{M}(\pi(I_j), I_j)
\end{equation}
where $\pi : I \rightarrow \Lambda$ is a selector that selects an algorithm $\Lambda_i \in \Lambda$ for instance $I_j \in I$ to maximize the overall performance metric $\mathcal{M}: \Lambda \times I \rightarrow \mathbb{R}$. 

Pertaining to the DAS setting proposed in this paper, time steps are introduced and the objective becomes seeking
\begin{equation}
    \pi^* = \mathop{\arg\max}\limits_{\pi \in \Pi} \sum_{j=1}^{M} \sum_{t=1}^{T} \mathcal{M}(\pi(I_j,t),I_j)
\end{equation}
where $\pi: I \times T \rightarrow \Lambda$ indicates that the selector should choose a proper algorithm at each optimization interval for each instance. Note that if we 1) define a reward function based on the metric $\mathcal{M}$, 2) consider choosing candidate algorithms at different time steps as actions, and 3) formulate the optimization state at each time step as states, the DAS problem can be regarded as an MDP, which could be solved by reinforcement learning methods such as the PPO~\cite{schulman2017proximal}.

\subsection{RL-DAS Overview}\label{Sec:overview}
According to the definition of DAS, the environment of DAS includes an algorithm pool, a problem set, and a population for optimization. To formulate the MDP, we design the \textit{state} to contain sufficient information through analysis of the fitness landscape and optimization status, the \textit{action} to select one algorithm from the algorithm pool, and the \textit{reward} to reflect the improvement of metric $\mathcal{M}$.
  The workflow of RL-DAS is presented in Algorithm~\ref{alg:pseudo}. 
The training on each problem instance starts with an initial population $P_0$ and terminates when either the maximum number of function evaluations $MaxFEs$ is exhausted or the best cost so far in the population is lower than the termination error. In the $t$-th step, information from the population $P_t$, the optimization problem and the algorithm pool $\Lambda$ is extracted and constructed as a state $s_t$. Based on the state, the policy agent $\pi_\theta$ decides an action $a_t$ that selects the next algorithm to switch. 
Additionally, we calculate the log-likelihood of the action, $\log\pi_\theta(a_t | s_t)$ which is used in PPO. With parameters inherited from a context memory $\Gamma$, an algorithm is warm-started from its last break and applied to envolve the population for a period of time (until the next decision step). After the period, the reward $r_t$ is observed, and a new population $P_{t+1}$ is produced, generating the new state information $s_{t+1}$. The updated contextual information of the algorithm is appended to the memory $\Gamma$ to prepare for the next restoration. The trajectories of states, actions, and rewards are recorded and then used by the PPO method to train the policy net $\pi_\theta$ and the critic net $v_\phi$. 
To fully utilize the recorded trajectories and promote the training efficiency, PPO updates the networks $K$ times after the completion of optimization. A larger $K$ leads to slower training and an elevated risk of overfitting on the current problem, whereas a smaller $K$ may restrict the utilization and the overall efficiency. In this paper, we set $K = 0.3\times T$.

In the following of this section, we introduce the design of MDP components including the state, action and reward in Section~\ref{Sec:samp}. The neural network design and the training process will be presented in Section~\ref{Sec:net}. Last, we introduce the aforementioned algorithm context memory mechanism to support the warm-start of different algorithms in Section~\ref{Sec:context}.

\begin{algorithm}[t]\footnotesize
\caption{Pseudo Code of RL-DAS}\label{alg:pseudo}
\KwIn{Policy $\pi_\theta$, Critic $v_\phi$, Instance Set $I$, Algorithm Pool $\Lambda$ with Context Memory $\Gamma$}
\KwOut{Trained Policy $\pi_\theta$, Critic $v_\phi$}
\For{$epoch \gets 0$ \KwTo $Epoch$}{
    \For{$instance \in I$}{
        
        Population $ P_0 \gets InitPopulation() $\;
        Initialize $\Gamma$\;
        $ t \gets 0 $\;
        \While{the termination condition is not met}{
            $ s_t \gets \text{State}(P_t, instance, \Lambda) $\;
            $ a_t, \log\pi_\theta(a_t | s_t) \gets \pi_\theta(a_t |s_t) $\;
            $ algorithm \gets \Lambda[a_t] $\;
            $ algorithm \gets \text{Restoration}(\Gamma, algorithm) $\;
            $ P_{t+1} \gets \text{Step}(algorithm, P_t, instance) $\;
            $ s_{t+1} \gets \text{State}(P_{t+1}, instance, \Lambda) $\;
                Get reward $adc_t$ by Eq.~(\ref{eq:adc})\;
                Get $<s_t, a_t, s_{t+1}, adc_t, \log\pi_\theta(a_t | s_t)>$\;
            $ t \gets t + 1 $\;
            Update $\Gamma$\;
        }
        Update $adc_{0:T}$ into $r_{0:T}$ by Eq.~(\ref{eq:reward})\;
        
        \For{$k \gets 1$ \KwTo $K$}{
            Update $\pi_\theta$ and $v_\phi$ by \textbf{PPO} method\;
        }
    } 
} 

\end{algorithm}

\subsection{MDP Components}\label{Sec:samp}

This section presents a detailed design of the MDP, including state, action, and reward, and explains how they enable the interaction between the environment and the DRL agent.



\subsubsection{State}

The state space of RL-DAS consists of Landscape Analysis (LA)
and Algorithm History (AH) information. For the LA part, nine scalar features ($feature_1, \cdots, feature_9$) are extracted by analyzing the population information and the random walk sampling information on the problem landscape, which forms a LA vector $f_\text{LA} \in \mathbb{R}^9$. 
The sampling information probes the fitness landscape, revealing the complexity and difficulty of the optimization problem. These features also reflect the interaction between the current population and the optimization problem. 
 Below we briefly introduce the LA features ($feature_1, \cdots, feature_9$), deferring the detailed formulations in Section 1 of our supplementary document.

\begin{itemize}
    \item $feature_1$: The cost value is a pivotal indicator of optimization status, making it our first feature.
   
    \item $feature_2$: The fitness distance correlation (FDC)~\cite{tomassini2005study} describes problem complexity by assessing the relationship between fitness values and solution distances. 
    \item $feature_3$: The dispersion difference~\cite{lunacek2006dispersion} measures the distribution difference between the top 10\% and the whole population.  
    It is applied to analyze the funnelity of the problem landscape: a single funnel problem has a smaller dispersion difference value as the top 10\% individuals gather around the optimal point and have smaller inner distance, otherwise for the multi-funnel landscape, this value would be much larger.
    \item $feature_4$: The maximum distance among population 
    describes the convergence state of population. 
    \item $feature_{5,6,7,8}$: These features gauge the evolvability of the population on the given problem. First, the negative slope coefficient (NSC)~\cite{vanneschi2004fitness} measures the difference between the cost change incurred by the population evolution and by making small tentative random walk steps. 
    The average neutral ratio (ANR)~\cite{vanneschi2007comprehensive} has a similar consideration of NSC, but it measures the performance changes between the current population and different sampled populations. 
    Inspired by Wang et al.~\cite{wang2017population}, we also propose the Best-Worst Improvement to measure the evolvability. The Best Improvement feature measures the optimization difficulty of the current population using the ratio of sampled individuals that make no improvement over the current population, while the Worst Improvement feature describes the optimization potential by the ratio of individuals that do not get worse.
    \item $feature_9$: 
    To keep the agent informed about computational budget consumption, we introduce the ratio of consumed $FEs$ over the maximum as a feature.
\end{itemize}
 Note that the function evaluations costed in the above process for state representation should be included in evaluating the termination condition.

As our environment also contains the algorithm pool, to 
provide the RL agent an additional contextual knowledge about the optimization ability of the candidate algorithms, we derive the AH information to characterize the optimization status from an algorithmic perspective. 
Suppose that there are $L$ algorithms in $\Lambda$. We maintain two shift vectors for each algorithm to characterize the population distribution change incurred by it. Specifically, let $SV_\text{best}^l \in \mathbb{R}^D$ and $SV_\text{worst}^l \in \mathbb{R}^D$ denote the average shift of the best and the worst individual caused by the algorithm $\Lambda^l$ in the previous optimization steps. 
 For a time interval $i$, if $\Lambda^l$ is selected to execute, it transforms the population $P$ with the best individual $x_\text{best}$ to a new population $P'$ with $x'_\text{best}$. Then, the shift vector of the best individual at the $i$th step, $sv_{i, \text{best}}^l \in \mathbb{R}^D$, is defined as:
\begin{equation}\label{eq:sv}
    sv_{i, \text{best}}^l = x'_\text{best} - x_\text{best}
\end{equation}
which presents the effect of $\Lambda^l$ in this step. Then, the $SV_\text{best}^l$ is formulated as:
\begin{equation}\label{eq:ah}
    SV_\text{best}^l = \frac{1}{H^l}\sum_{i=1}^{H^l}{sv_{i, \text{best}}^l}
\end{equation}
where $H^l$ is the number of intervals that the algorithm $\Lambda^l$ is selected. The calculation of $SV_\text{worst}^l$ is similar. Notably, in the case that the algorithm $\Lambda^l$ is never selected by the RL agent ($H^l = 0$), its $SV_\text{best}^l$ and $SV_\text{worst}^l$ are zero vectors. 

Based on the above, $f_\text{AH}$ is a collection of the $L\times 2$ shifting vectors of all candidate algorithm, which captures the behavior histories of the candidate algorithms and helps RL agent learn about the characteristics of candidate algorithms. 
\begin{equation}\label{eq:aoh}
    f_\text{AH} = \left[\{SV_\text{best}^1,SV_\text{worst}^1\}\cdots\{SV_\text{best}^L,SV_\text{worst}^L\}\right]
\end{equation}

Finally, the complete state in the MDP of RL-DAS is the integration of the above LA and AH information.
\begin{equation}\label{eq:state}
    state = \{ f_\text{LA} \in \mathbb{R}^9, f_\text{AH} \in \mathbb{R}^{2L \times D} \}
\end{equation}
An intuitive explanation of this state design is to provide the RL agent with awareness of both the problem instance characteristics and the historical performance of algorithms. This enables the RL agent to make more informed decisions when selecting algorithms during the optimization process.

\subsubsection{Action}

Since the motivation behind RL-DAS is to periodically switch optimization algorithms to achieve complementary performance, the action is designed as an integer number indicating the index of the selected algorithm in a pool containing $L$ candidate algorithms, denoted as $a \in [1, L]$.

\subsubsection{Reward}

In order to guide the agent towards achieving a lower evaluation cost, the reward function should consider the absolute cost reduction at each time step $t$:
\begin{equation}\label{eq:adc}
    adc_t = \frac{cost^*_{t-1} - cost^*_{t}}{cost^*_0}
\end{equation}
where $cost^*_t$ and $cost^*_0$ are the best cost until the $t$-th step and the best cost in the initial population, respectively (where $cost^*_0$ serves as a normalization factor). This measures the performance improvement brought in the $t$-th step optimization. 

In addition, when employing EC algorithms for optimization, we usually pay attention not only to the descent of cost but also to the speed of convergence. The latter can be measured by the number of consumed $FEs$ to the termination condition. 
Therefore, our reward function is defined as an adapted absolute descent of cost, by 
introducing the measurement of optimization speed:
\begin{equation}\label{eq:reward}
    r_t = \frac{cost^*_{t-1} - cost^*_{t}}{cost^*_0} \times \frac{MaxFEs}{FEs_\text{end}}
\end{equation}
where 
$FEs_\text{end}$ is the number of consumed FEs in reaching the termination condition. 
The agent is encouraged to search for an optimal strategy capable of finding the solution with the lowest cost, while making the most efficient use of resources.


\subsection{Network Design}\label{Sec:net}

\begin{figure}[!t]
\centering
\includegraphics[width=0.65\columnwidth]{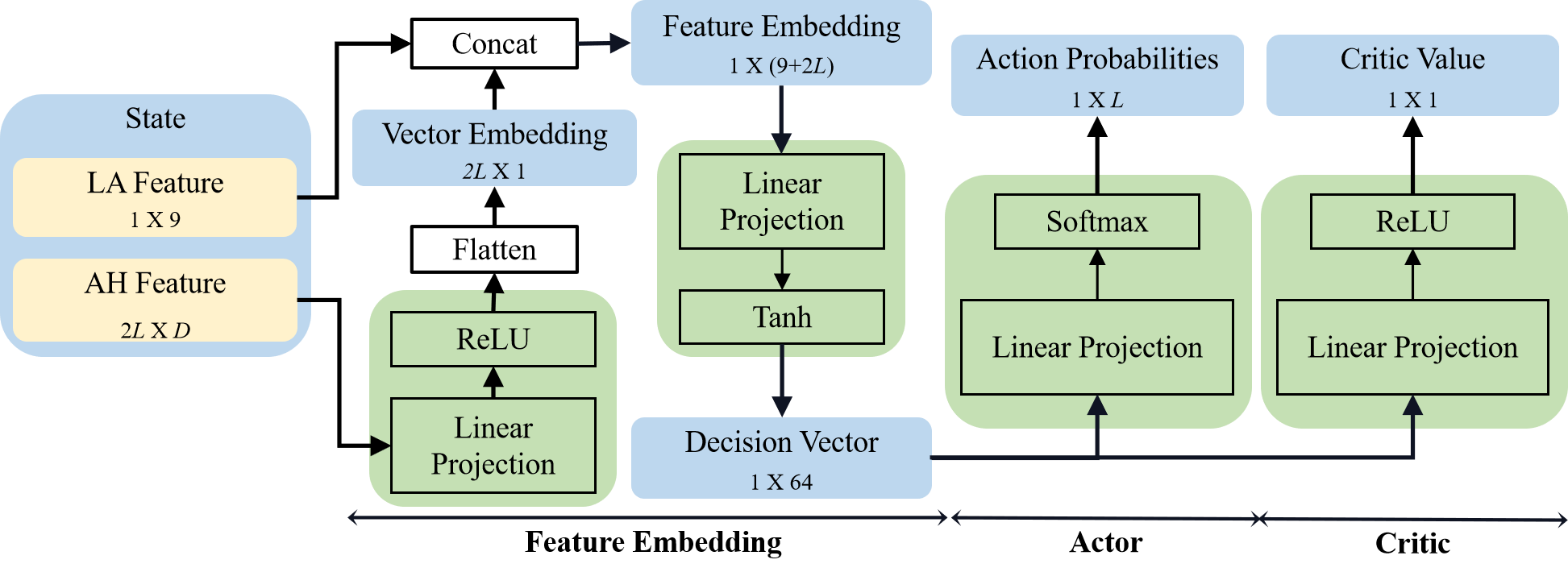}
\caption{The Neural Network workflow for $\pi_\theta$ (Actor) and $v_\phi$ (Critic).}
\label{net}
\end{figure}
As shown in Fig.~\ref{net}, the network is composed of three modules: Feature Embedding, Actor, and Critic. The AH feature and the LA feature are first fused to form the state representation vector. Based on the representation, the Actor outputs the probability distribution over the candidate algorithms, and the Critic estimates a return value. In the implementation of RL-DAS, we feed the network a batch of problems simultaneously to enhance the convergence of training. In the subsequent description, we omit the time step subscript $t$ and the batch dimensions for better readability.
\subsubsection{Feature Embedding}
The Feature Embedding module fuses the $f_\text{LA}$ and the $f_\text{AH}$. First, we feed the $f_\text{AH}$ into a two-layer feed forward network to generate a more compact embedding vector $V_\text{AH}$ as 
\begin{equation}
    V_\text{AH} = \sigma(W^{(2)}_\text{ve}\sigma(W^{(1)}_\text{ve}f_\text{AH} + b^{(1)}_\text{ve}) + b^{(2)}_\text{ve})
\end{equation}
where $W^{(1)}_\text{ve} \in \mathbb{R}^{D\times 64}, W^{(2)}_\text{ve} \in \mathbb{R}^{64\times 1}$ are two network layers and the activation function $\sigma$ is ReLU. Through the two layers, the AH information is embedded in a $2L$ compact vector, whose length is in the similar magnitude of the 9-dimensional $f_\text{LA}$. The two types of features are concatenated and then fused by another network layer to generate the decision vector representation as
\begin{equation}
    DV=\mathrm{Tanh}\left( W_{\mathrm{dv}}\left( f_{\mathrm{LA}}\oplus \mathrm{Flatten}\left( V_{\mathrm{AH}} \right) \right) +b_{\mathrm{dv}} \right) 
    \label{eq:dv}
\end{equation}
where the weight for the decision vector $W_\text{dv} \in \mathbb{R}^{(9+2L)\times 64}$, and the activation function Tanh is applied to limit the possible extreme values in the features.
\subsubsection{Actor} 
Actor decides the probability for selecting an algorithm from the candidate algorithm pool. It first maps $DV$ in Eq.~(\ref{eq:dv}) to a logits vector through a two-layer feed-forward network. Then, after the $\text{Softmax}$ operation, the actor outputs the probability distribution on the candidate algorithm pool, which are then used to sample an algorithm:
\begin{equation}
    \pi = \text{Softmax}(W^{(2)}_\text{actor}\text{Tanh}(W^{(1)}_\text{actor}DV + b^{(1)}_\text{actor}) + b^{(2)}_\text{actor})
\end{equation}
where $W^{(1)}_\text{actor} \in \mathbb{R}^{64\times 16}$ and $W^{(2)}_\text{actor} \in \mathbb{R}^{16\times L}$. The policy $\pi$ indicates the probabilities of each algorithm to be selected.

\subsubsection{Critic}
Critic feeds $DV$ into a two-layer feed forward network to obtain the estimated return value of the state.
\begin{equation}
    b = \sigma(W^{(2)}_\text{critic}\sigma(DV^\top W^{(1)}_\text{critic} + b^{(1)}_\text{critic}) + b^{(2)}_\text{critic})
\end{equation}
where $W^{(1)}_\text{critic} \in \mathbb{R}^{D\times 64}$ and $W^{(2)}_\text{critic} \in \mathbb{R}^{64\times 1}$ are two network layers, and the activation function $\sigma$ in both layers is ReLU.

\subsection{Algorithm Context Memory}\label{Sec:context}

Modern EC algorithms usually contain some contextual information, such as the dynamically determined parameters, some statistical measures or indicators, and other configurations during the optimization. 
For single algorithm execution, such information is smoothly updated along the optimization progress. However, the DAS studied in this paper may break the smoothness of updating the above information when switching algorithms since different algorithms consider different contextual information. This results in a \emph{warm up} issue, i.e., how to properly manage the context information of different algorithms and then perform context restoration for a smooth search.
To address this issue, we first separate the algorithm-specific contexts from the common contexts among the candidate algorithms.
Then, we design a nested dictionary-like context memory to manage them:
\begin{equation}
    \Gamma := \begin{cases}
        \Lambda_1: \{\text{ctx}_{1,1}, \text{ctx}_{1,2}, \cdots\} \\
        \indent \vdots \\
        \Lambda_L: \{\text{ctx}_{L,1}, \text{ctx}_{L,2}, \cdots\} \\
        Common: \{\text{ctx}_1, \text{ctx}_2, \cdots\} \\
    \end{cases}
\end{equation}

Here, $\text{ctx}_{i,*}$ indicates contexts for the $i$-th algorithm, which is updated when the corresponding algorithm is selected for execution. The $Common$ dictionary contains some common contexts among the candidate algorithms,
which can be updated by multiple candidates.  When an algorithm $\Lambda_i$ needs to be restored, its dictionary $\Gamma[\Lambda_i]$ is indexed, and the contexts inside $\Gamma[\Lambda_i]$ and $Common$ are restored, in order to support the warm-start of $\Lambda_i$. 
This restoration method can be generalized to various algorithms
, as it does not propose any special requirements on the contexts.

\section{RL-DAS on DE: A Proof-of-Principle Study}\label{Sec:case}
The design of RL-DAS can be applied to various EC algorithms, including GA, PSO, DE, and their mixtures, each with corresponding algorithm pools. However, in this paper, for the sake of convenience in the comparison and discussion, we focus on presenting a proof-of-principle study using representative DE algorithms. 
As mentioned in Section~\ref{Sec:DE}, DE is a high-profile field in EC. Although exhibits powerful performance in continuous optimization, different DE variants generally exhibit different performances on different problem instances. 
We realize the above RL-DAS framework to dynamically schedule three DE algorithms along the optimization process, which is detailed in this section.


\subsection{Candidate Algorithm Pool}
We adopt three advanced DE algorithms from the CEC 2021 competition~\cite{cec2021TR}, namely, JDE21~\cite{brest2021self}, MadDE~\cite{biswas2021improving} and NL-SHADE-RSP~\cite{stanovov2021nl}.
These three algorithms have been introduced in Section~\ref{Sec:DE}, which have different strengths and weaknesses in terms of balancing the exploration and exploitation of the population. According to our experiments (which will be discussion in Section 3 of our supplementary document), these three algorithms present different search characteristics and exhibit complementary performance in different optimization stages of different problem instances, 
which leaves room for our RL-DAS agent to learn a desired collaboration between them. 

\subsection{Schedule Interval}\label{Sec:SI}

The interval between schedules plays a significant role in the training efficiency of DRL agents and the optimization performance of EC. On the one hand, a short interval allows the agent to switch among candidates more frequently and accumulate more transitions, helps the DRL agent to learn faster. However, on the other hand, frequent changes can break the integrity of algorithms, resulting in poor optimization performance because the algorithm can be switched before making any substantial improvement. 
A long interval allows algorithms to explore the search space sufficiently, but the agent learns less, and the flexibility of DAS is also decreased. An extreme case is that when the interval is as long as the complete optimization budget, the DAS degrades to the traditional AS. Therefore, determining the interval requires balancing the optimization performance and the training efficiency.

According to the RL practice, the common length of an RL trajectory ranges from $T_\text{min} = 10$ to $T_\text{max} = 200$~\cite{gym}. 
To this end, considering the optimization length in EC, which is usually controlled by the $MaxFEs$, the suggested schedule interval of RL-DAS is 
$\varDelta \in \left[ \frac{MaxFEs}{T_{\max}},\frac{MaxFEs}{T_{\min}} \right]
$. In essence, at every $\varDelta$ FEs, the framework attempts to insert a transition (algorithm switching) step of RL. In practical execution, the framework will wait until the current generation of the DE algorithm completed before starting the transition step. As a result, the real scheduling interval, $\tilde{\varDelta}$, may be equal to or slightly larger than the planned $\varDelta$.
In Section~\ref{Exp:AS}, we will conduct a grid experiment to investigate interval lengths 
that support good performance with acceptable training efficiency. 

\subsection{Algorithm Restoration}

The three algorithms in our candidate pool adopt many adaptive methods, which makes restoring these contexts a challenge. 
For JDE21, the contexts contain parameters such as the probabilities of adaptive mutation and crossover ($\tau_1$ and $\tau_2$), the parameters determine the count of stagnation ($ageLmt$, $eps$) and the controller of reinitialization ($myEqs$). MadDE adopts multiple operators, whose contexts involve the mutation probabilities ($p_m$ and $p_{best}$) and the crossover probability ($p_{qBX}$). In NL-SHADE-RSP, the probability of using archive ($p_A$) and its adaption parameter ($n_A$) are included. The successful parameters ($S_F$ and $S_{Cr}$) for JDE21 and the parameter memories ($M_F$ and $M_{Cr}$) for MadDE and NL-SHADE-RSP, as well as their elite archive ($Arch$) are stored in the $Common$. To summarize, the complete context memory $\Gamma$ for the three algorithms is organized in the following form:
\begin{equation}
    \Gamma := \begin{cases}
        \text{JDE21}: \{\tau_1, \tau_2, ageLmt, eps, myEqs\} \\
        \text{MadDE}: \{p_m, p_{best}, p_{qBX}\} \\
        \text{NL-SHADE-RSP}: \{n_A, p_A\} \\
        Common: \{S_F, S_{Cr}, M_F, M_{Cr}, Arch\} \\
    \end{cases}
\end{equation}
For example, when MadDE is selected, the algorithm finds the linked dictionary and uses $p_m$, which is updated and frozen since last time MadDE being selected, to sample mutation operators. At the end of each generation, the probabilities are updated according to the performance of mutations. When MadDE is switched (e.g., to use JDE21 or NL-SHADE-RSP), the adaptive contexts in $\Gamma$ are replaced with those updated $p_m$. 

\begin{figure}[!t]
\centering
\includegraphics[width=0.50\columnwidth]{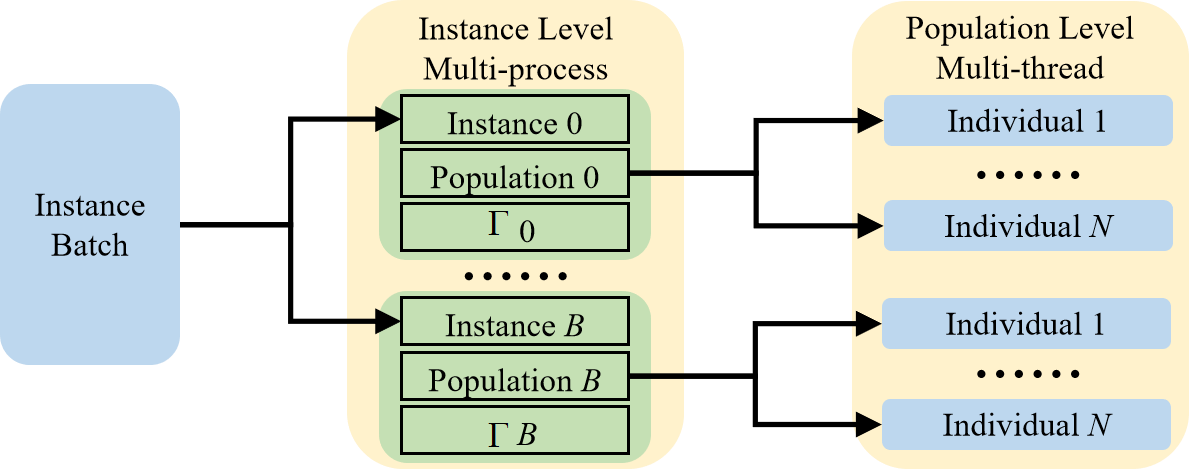}
\caption{The structure of the hierarchical parallel mechanism. $B$ indicates the batch size.}
\label{para}
\end{figure}

\subsection{Parallelization}

In RL-DAS, the training of the DRL agent is based on the sampled transitions from a period of EC iterations. Usually the agent requires a lot of sampled transitions to converge, which would consume much computational time. 
To address this low training efficiency problem, we adopt a hierarchical parallel mechanism that accelerates the sampling at both the instance level and the population level.

Pertaining to the instance level, we use a parallel method to sample transitions from a problem instance batch. First, a batch of problem instances is sampled from the problem suite. Accordingly, we generate the same number of populations and algorithm architectures. These instances, populations, and algorithms are then run in parallel with sub-processes. The agent processes their features in batch and returns the actions to their corresponding sub-processes, which determines the next algorithm to use on each instance. Note that the populations and the algorithm context memories are not shared among sub-processes. Pertaining to the population level, we design operators to process the entire population at once using the multi-thread parallel method from Numpy, instead of handling individuals one by one.

\section{Experiment}\label{Sec:Exp}
Our experiments study the following research questions:
\begin{itemize}
\setlength{\itemsep}{0pt}
\setlength{\parsep}{0pt}
\setlength{\parskip}{0pt}
    \item RQ1: 
    How does the performance of our RL-DAS compare with its backbone candidate algorithms, an optimal estimate of traditional AS, and other advanced DE variants (including adaptive, ensemble, and RL-assisted ones)?  
    See Section~\ref{Exp:PC}.
    \item RQ2: How well does RL-DAS generalize to unseen problem instances? See Section~\ref{Exp:ZT}.
    \item  RQ3: What are the suggested configurations for RL-DAS? See Section~\ref{Exp:AS}.
\end{itemize}

\subsection{Benchmark Generation}\label{Exp:DG}
There are different benchmark test suites proposed in the past two decades to test the performance of EC algorithms, which usually contain 10-20 problem instances. For example, the CEC2021 Competition on Single Objective BBO~\cite{cec2021TR} contains 
1 unimodal problem (F1), 3 basic problems (F2 - F4), 3 Hybrid problems (F5 - F7) and 3 Composition problems (F8 - F10).  For a few existing RL-assisted EC~\cite{dedqn,deddqn}, they train their agents on a small number of benchmark problem instances and also test on a small dataset. Though shown improved performance, 
their learned models may overfit to these fixed instances and lose the generalization on more problem instances. 
Meanwhile, since the training and test samples are limited, this further hinders their final performance.
To this end, this paper uses an augmented CEC2021 benchmark that randomly generates a class of instances for each benchmark problem in the CEC2021 (denoted as C1 - C10) and requires the agent to achieve generalization performance on the class of instances. The instances in a class are similar problems but use different shifting vectors and rotation matrices. Benefited from the exquisite design of CEC problems, all instances have known optima. Additionally, we adopt a mix-class instance set that contains instances from all problem classes (denoted as C11) to further inspect the generalization ability. Specifically, following the Technical Report~\cite{cec2021TR}, we generate rotation matrices with the Gram-Schmidt method~\cite{hilbert1989theorie} and uniformly sample each dimension of the shifting vector in the range [-80, 80]. The other configurations follow the original settings in CEC2021, such as the selection of subproblems of Hybrid and Composition problems. 
In this way, we expand each original benchmark into a problem class composed of 2048 instances. The $K$-fold cross-validation method is adopted to separate instances into training sets and validation sets, where we use $K\!=\!4$, meaning that 512 instances will be taken as validations. Furthermore, another 2048 instances are generated for each problem class to test the performance. 

\subsection{Competitors}

In this paper, we apply a DRL agent to dynamically select an optimization algorithm at each time step from the three algorithms in the CEC2021 competition: JDE21~\cite{brest2021self}, NL-SHADE-RSP~\cite{stanovov2021nl}, and MadDE~\cite{biswas2021improving}. First, we compare our RL-DAS algorithm with these three algorithms, as well as with a random selection baseline (Rand-DAS), where the RL agent in RL-DAS is replaced with random selection.
Second, to present the difference between DAS and the traditional AS, we compare it with an AS$^*$ algorithm that possesses the performance upper bound of AS. 
Third, to further validate the effectiveness of RL-DAS, we compare it with some other advanced DEs, including an adaptive DE named EPSDE~\cite{mallipeddi2011differential}, an ensemble DE named EDEV~\cite{wu2018ensemble}, and 
three RL-assisted DEs, namely DEDQN~\cite{dedqn}, DEDDQN~\cite{deddqn} and LDE~\cite{lde}. 
All algorithms are implemented in Python. The  control parameters and more experiment settings are provided in Section~2 of the supplementary document.

\subsection{Performance Comparisons}
\label{Exp:PC}

We first compare the optimization results and the optimization speed on 10D and 20D augmented CEC2021 benchmarks. Note that, for all comparisons in this paper, we report the cost decent in percentage, and we additionally provide the absolute cost values in Section 4 of our supplementary document. 

\begin{table*}[]
\centering
\caption{
Comparing RL-DAS with comparison algorithms on Shifted \& Rotated 10D problems. }
\resizebox{0.98\textwidth}{!}{%
\begin{tabular}{cc|c|lllllllllll|c}
\hline
  \multicolumn{2}{c|}{\multirow{3}{*}{Taxonomy}} 
& \multicolumn{1}{c|}{\multirow{3}{*}{Algorithm}} 
& \multicolumn{11}{c|}{\multirow{1}{*}{Problem Class}}
& \multicolumn{1}{c}{\multirow{3}{*}{\begin{tabular}[c]{@{}c@{}}Significance\\Test\\(Win/Tie/Loss)\end{tabular}}}
\\ \cline{4-14}
 & &
 & \multicolumn{1}{l}{\multirow{2}{*}{\quad C1}}
 & \multicolumn{1}{l}{\multirow{2}{*}{\quad C2}}
 & \multicolumn{1}{l}{\multirow{2}{*}{\quad C3}}
 & \multicolumn{1}{l}{\multirow{2}{*}{\quad C4}}
 & \multicolumn{1}{l}{\multirow{2}{*}{\quad C5}}
 & \multicolumn{1}{l}{\multirow{2}{*}{\quad C6}}
 & \multicolumn{1}{l}{\multirow{2}{*}{\quad C7}}
 & \multicolumn{1}{l}{\multirow{2}{*}{\quad C8}}
 & \multicolumn{1}{l}{\multirow{2}{*}{\quad C9}}
 & \multicolumn{1}{l}{\multirow{2}{*}{\quad C10}}
 & \multicolumn{1}{l|}{\multirow{2}{*}{\quad C11}}
 & 
 \\
 & & & & & & & & & & & & & &
 \\ \hline
\multicolumn{2}{c|}{\multirow{3}{*}{DAS}}
& \textbf{RL-DAS}
& \begin{tabular}[c]{@{}c@{}}\textbf{100.0}\%\\ \textbf{1.127e5}\end{tabular}
& \begin{tabular}[c]{@{}c@{}}\textbf{85.97}\%\\ \textbf{2.000e5}\end{tabular}
& \begin{tabular}[c]{@{}c@{}}\textbf{95.63}\%\\ \textbf{2.000e5}\end{tabular}
& \begin{tabular}[c]{@{}c@{}}\textbf{100.0}\%\\ 2.000e5\end{tabular}
& \begin{tabular}[c]{@{}c@{}}\textbf{100.0}\%\\ 1.996e5\end{tabular}
& \begin{tabular}[c]{@{}c@{}}\textbf{99.39}\%\\ \textbf{2.000e5}\end{tabular}
& \begin{tabular}[c]{@{}c@{}}\textbf{100.0}\%\\ \textbf{2.000e5}\end{tabular}
& \begin{tabular}[c]{@{}c@{}}\textbf{80.48}\%\\ 1.821e5\end{tabular}
& \begin{tabular}[c]{@{}c@{}}\textbf{90.22}\%\\ \textbf{1.882e5}\end{tabular}
& \begin{tabular}[c]{@{}c@{}}88.79\%\\ \textbf{1.914e5}\end{tabular}
& \begin{tabular}[c]{@{}c@{}}\textbf{93.94}\%\\ 1.903e5\end{tabular}
& - / - / -
\\
&
& \textbf{Rand-DAS}
& \begin{tabular}[c]{@{}c@{}}\textbf{100.0}\% $\approx$\\ 1.359e5 $-$\end{tabular}
& \begin{tabular}[c]{@{}c@{}}82.87\% $-$\\ \textbf{2.000e5} $\approx$\end{tabular}
& \begin{tabular}[c]{@{}c@{}}95.40\% $-$\\ \textbf{2.000e5} $\approx$\end{tabular}
& \begin{tabular}[c]{@{}c@{}}\textbf{100.0}\% $\approx$\\ \textbf{1.998e5} $\approx$\end{tabular}
& \begin{tabular}[c]{@{}c@{}}\textbf{100.0}\% $\approx$\\ 1.994e5 $\approx$\end{tabular}
& \begin{tabular}[c]{@{}c@{}}99.33\% $\approx$\\ \textbf{2.000e5} $\approx$\end{tabular}
& \begin{tabular}[c]{@{}c@{}}\textbf{100.0}\% $\approx$\\ \textbf{2.000e5} $\approx$\end{tabular}
& \begin{tabular}[c]{@{}c@{}}79.87\% $-$\\ \textbf{1.795e5} $+$\end{tabular}
& \begin{tabular}[c]{@{}c@{}}89.91\% $-$\\ 1.885e5 $-$\end{tabular}
& \begin{tabular}[c]{@{}c@{}}88.52\% $-$\\ 1.915e5 $\approx$\end{tabular}
& \begin{tabular}[c]{@{}c@{}}93.69\% $-$\\ 1.891e5 $+$\end{tabular}
& 7 / 4 / 0
\\ \hline
\multicolumn{2}{c|}{\multirow{1}{*}{AS}}
& \textbf{AS$^*$}
& \begin{tabular}[c]{@{}c@{}}\textbf{100.0}\% $\approx$\\1.139e5 $-$\end{tabular}
& \begin{tabular}[c]{@{}c@{}}85.63\% $-$\\ \textbf{2.000e5} $\approx$\end{tabular}
& \begin{tabular}[c]{@{}c@{}}95.40\% $-$\\ \textbf{2.000e5} $\approx$\end{tabular}
& \begin{tabular}[c]{@{}c@{}}\textbf{100.0}\% $\approx$\\ 2.000e5 $\approx$\end{tabular}
& \begin{tabular}[c]{@{}c@{}}\textbf{100.0}\% $\approx$\\ \textbf{1.945e5} $+$\end{tabular}
& \begin{tabular}[c]{@{}c@{}}99.34\% $\approx$\\ \textbf{2.000e5} $\approx$\end{tabular}
& \begin{tabular}[c]{@{}c@{}}\textbf{100.0}\% $\approx$\\ \textbf{2.000e5} $\approx$\end{tabular}
& \begin{tabular}[c]{@{}c@{}}79.86\% $-$\\ 1.865e5 $-$\end{tabular}
& \begin{tabular}[c]{@{}c@{}}89.94\% $-$\\ 1.884e5 $\approx$\end{tabular}
& \begin{tabular}[c]{@{}c@{}}88.55\% $-$\\ 1.917e5 $\approx$\end{tabular}
& \begin{tabular}[c]{@{}c@{}}93.71\% $-$\\ \textbf{1.870e5} $+$\end{tabular}
& 7 / 3 / 1
\\ \hline
\multicolumn{2}{c|}{\multirow{5}{*}{Backbones}}
& \textbf{JDE21}
& \begin{tabular}[c]{@{}c@{}}\textbf{100.0}\% $\approx$\\ 1.914e5 $-$\end{tabular}
& \begin{tabular}[c]{@{}c@{}}73.52\% $-$\\ \textbf{2.000e5} $\approx$\end{tabular}
& \begin{tabular}[c]{@{}c@{}}94.97\% $-$\\ \textbf{2.000e5} $\approx$\end{tabular}
& \begin{tabular}[c]{@{}c@{}}\textbf{100.0}\% $\approx$\\ 2.000e5 $\approx$\end{tabular}
& \begin{tabular}[c]{@{}c@{}}99.99\% $-$\\ 2.000e5 $-$\end{tabular}
& \begin{tabular}[c]{@{}c@{}}99.01\% $-$\\ \textbf{2.000e5} $\approx$\end{tabular}
& \begin{tabular}[c]{@{}c@{}}99.99\% $\approx$\\ \textbf{2.000e5} $\approx$\end{tabular}
& \begin{tabular}[c]{@{}c@{}}78.76\% $-$\\ 1.935e5 $-$\end{tabular}
& \begin{tabular}[c]{@{}c@{}}88.47\% $-$\\ 1.954e5 $-$\end{tabular}
& \begin{tabular}[c]{@{}c@{}}88.43\% $-$\\ 1.954e5 $-$\end{tabular}
& \begin{tabular}[c]{@{}c@{}}92.02\% $-$\\ 1.975e5 $-$\end{tabular}
& 9 / 2 / 0
\\
&
& \textbf{MadDE}
& \begin{tabular}[c]{@{}c@{}}\textbf{100.0}\% $\approx$\\ 1.160e5 $-$\end{tabular}
& \begin{tabular}[c]{@{}c@{}}83.64\% $-$\\ \textbf{2.000e5} $\approx$\end{tabular}
& \begin{tabular}[c]{@{}c@{}}95.39\% $-$\\ \textbf{2.000e5} $\approx$\end{tabular}
& \begin{tabular}[c]{@{}c@{}}\textbf{100.0}\% $\approx$\\ 2.000e5 $\approx$\end{tabular}
& \begin{tabular}[c]{@{}c@{}}\textbf{100.0}\% $\approx$\\ 1.955e5 $+$\end{tabular}
& \begin{tabular}[c]{@{}c@{}}99.34\% $\approx$\\ \textbf{2.000e5} $\approx$\end{tabular}
& \begin{tabular}[c]{@{}c@{}}\textbf{100.0}\% $\approx$\\ \textbf{2.000e5} $\approx$\end{tabular}
& \begin{tabular}[c]{@{}c@{}}79.03\% $-$\\ 1.867e5 $-$\end{tabular}
& \begin{tabular}[c]{@{}c@{}}89.77\% $-$\\ 1.889e5 $-$\end{tabular}
& \begin{tabular}[c]{@{}c@{}}87.93\% $-$\\ 1.920e5 $-$\end{tabular}
& \begin{tabular}[c]{@{}c@{}}93.71\% $-$\\ 1.878e5 $+$\end{tabular}
& 7 / 3 / 1
\\
&
& \textbf{NL-SHADE-RSP}
& \begin{tabular}[c]{@{}c@{}}{100.0}\% $\approx$\\ {2.000e5} $-$\end{tabular}
& \begin{tabular}[c]{@{}c@{}}80.74\% $-$\\ \textbf{2.000e5} $\approx$\end{tabular}
& \begin{tabular}[c]{@{}c@{}}95.22\% $-$\\ \textbf{2.000e5} $\approx$\end{tabular}
& \begin{tabular}[c]{@{}c@{}}\textbf{100.0}\% $\approx$\\ 2.000e5 $\approx$\end{tabular}
& \begin{tabular}[c]{@{}c@{}}\textbf{100.0}\% $\approx$\\ 2.000e5 $-$\end{tabular}
& \begin{tabular}[c]{@{}c@{}}97.74\% $-$\\ \textbf{2.000e5} $\approx$\end{tabular}
& \begin{tabular}[c]{@{}c@{}}\textbf{100.0}\% $\approx$\\ \textbf{2.000e5} $\approx$\end{tabular}
& \begin{tabular}[c]{@{}c@{}}79.41\% $-$\\ 1.865e5 $-$\end{tabular}
& \begin{tabular}[c]{@{}c@{}}89.89\% $-$\\ 1.884e5 $-$\end{tabular}
& \begin{tabular}[c]{@{}c@{}}88.46\% $-$\\ 1.917e5 $\approx$\end{tabular}
& \begin{tabular}[c]{@{}c@{}}92.86\% $-$\\ 1.990e5 $-$\end{tabular}
& 9 / 2 / 0
\\ \hline
\multicolumn{1}{c}{\multirow{9}{*}{\begin{tabular}[c]{@{}c@{}}Other\\Advanced\\DE\\Variants\end{tabular}}}
& \multicolumn{1}{|c|}{\multirow{1}{*}{Adaptive}}
& \textbf{EPSDE}
& \begin{tabular}[c]{@{}c@{}}\textbf{100.0}\% $\approx$\\ 1.476e5 $-$\end{tabular}
& \begin{tabular}[c]{@{}c@{}}75.73\% $-$\\ \textbf{2.000e5} $\approx$\end{tabular}
& \begin{tabular}[c]{@{}c@{}}95.48\% $\approx$\\ \textbf{2.000e5} $\approx$\end{tabular}
& \begin{tabular}[c]{@{}c@{}}\textbf{100.0}\% $\approx$\\ 2.000e5 $\approx$\end{tabular}
& \begin{tabular}[c]{@{}c@{}}\textbf{100.0}\% $\approx$\\ 1.967e5 $-$\end{tabular}
& \begin{tabular}[c]{@{}c@{}}99.23\% $-$\\ \textbf{2.000e5} $\approx$\end{tabular}
& \begin{tabular}[c]{@{}c@{}}\textbf{100.0}\% $\approx$\\ \textbf{2.000e5} $\approx$\end{tabular}
& \begin{tabular}[c]{@{}c@{}}76.44\% $-$\\ 1.834e5 $-$\end{tabular}
& \begin{tabular}[c]{@{}c@{}}87.21\% $-$\\ 1.896e5 $-$\end{tabular}
& \begin{tabular}[c]{@{}c@{}}90.18\% $+$\\ 1.983e5 $-$\end{tabular}
& \begin{tabular}[c]{@{}c@{}}92.43\% $-$\\ 1.968e5 $-$\end{tabular}
& 7 / 3 / 1
\\ \cline{2-15}
& \multicolumn{1}{|c|}{\multirow{1}{*}{Ensemble}}
& \textbf{EDEV}
& \begin{tabular}[c]{@{}c@{}}\textbf{100.0}\% $\approx$\\ 1.402e5 $-$\end{tabular}
& \begin{tabular}[c]{@{}c@{}}77.45\% $-$\\ \textbf{2.000e5} $\approx$\end{tabular}
& \begin{tabular}[c]{@{}c@{}}95.53\% $-$\\ \textbf{2.000e5} $\approx$\end{tabular}
& \begin{tabular}[c]{@{}c@{}}\textbf{100.0}\% $\approx$\\ {2.000e5} $\approx$\end{tabular}
& \begin{tabular}[c]{@{}c@{}}\textbf{100.0}\% $\approx$\\ 1.960e5 $-$\end{tabular}
& \begin{tabular}[c]{@{}c@{}}99.27\% $-$\\ \textbf{2.000e5} $\approx$\end{tabular}
& \begin{tabular}[c]{@{}c@{}}\textbf{100.0}\% $\approx$\\ \textbf{2.000e5} $\approx$\end{tabular}
& \begin{tabular}[c]{@{}c@{}}77.37\% $-$\\ 1.827e5 $-$\end{tabular}
& \begin{tabular}[c]{@{}c@{}}87.65\% $-$\\ 1.889e5 $-$\end{tabular}
& \begin{tabular}[c]{@{}c@{}}\textbf{91.06}\% $+$\\ 1.980e5 $-$\end{tabular}
& \begin{tabular}[c]{@{}c@{}}93.02\% $-$\\ 1.909e5 $-$\end{tabular}
& 7 / 3 / 1
\\ \cline{2-15} 
& \multicolumn{1}{|c|}{\multirow{5}{*}{\begin{tabular}[c]{@{}c@{}}RL\\Assisted\end{tabular}}}
& \textbf{DEDQN}
& \begin{tabular}[c]{@{}c@{}}100.0\% $-$\\ 1.964e5 $-$\end{tabular}
& \begin{tabular}[c]{@{}c@{}}71.65\% $-$\\ \textbf{2.000e5} $\approx$\end{tabular}
& \begin{tabular}[c]{@{}c@{}}92.43\% $-$\\ \textbf{2.000e5} $\approx$\end{tabular}
& \begin{tabular}[c]{@{}c@{}}\textbf{100.0}\% $\approx$\\ 2.000e5 $\approx$\end{tabular}
& \begin{tabular}[c]{@{}c@{}}\textbf{99.99}\% $-$\\ 2.000e5 $-$\end{tabular}
& \begin{tabular}[c]{@{}c@{}}87.03\% $-$\\ \textbf{2.000e5} $\approx$\end{tabular}
& \begin{tabular}[c]{@{}c@{}}\textbf{100.0}\% $\approx$\\ \textbf{2.000e5} $\approx$\end{tabular}
& \begin{tabular}[c]{@{}c@{}}76.09\% $-$\\ 1.875e5 $-$\end{tabular}
& \begin{tabular}[c]{@{}c@{}}83.42\% $-$\\ 1.904e5 $-$\end{tabular}
& \begin{tabular}[c]{@{}c@{}}87.85\% $-$\\ 1.967e5 $-$\end{tabular}
& \begin{tabular}[c]{@{}c@{}}91.36\% $-$\\ 1.970e5 $-$\end{tabular}
& 9 / 2 / 0
\\
& \multicolumn{1}{|c|}{}
& \textbf{DEDDQN}
& \begin{tabular}[c]{@{}c@{}}100.0\% $-$\\ 1.989e5 $-$\end{tabular}
& \begin{tabular}[c]{@{}c@{}}71.98\% $-$\\ \textbf{2.000e5} $\approx$\end{tabular}
& \begin{tabular}[c]{@{}c@{}}93.23\% $-$\\ \textbf{2.000e5} $\approx$\end{tabular}
& \begin{tabular}[c]{@{}c@{}}\textbf{100.0}\% $\approx$\\ 2.000e5 $\approx$\end{tabular}
& \begin{tabular}[c]{@{}c@{}}\textbf{99.99}\% $-$\\ 2.000e5 $-$\end{tabular}
& \begin{tabular}[c]{@{}c@{}}88.77\% $-$\\ \textbf{2.000e5} $\approx$\end{tabular}
& \begin{tabular}[c]{@{}c@{}}\textbf{100.0}\% $\approx$\\ \textbf{2.000e5} $\approx$\end{tabular}
& \begin{tabular}[c]{@{}c@{}}76.85\% $-$\\ 1.864e5 $-$\end{tabular}
& \begin{tabular}[c]{@{}c@{}}84.52\% $-$\\ 1.890e5 $-$\end{tabular}
& \begin{tabular}[c]{@{}c@{}}88.07\% $-$\\ 1.956e5 $-$\end{tabular}
& \begin{tabular}[c]{@{}c@{}}91.26\% $-$\\ 1.969e5 $-$\end{tabular}
& 9 / 2 / 0
\\
& \multicolumn{1}{|c|}{}
& \textbf{LDE}
& \begin{tabular}[c]{@{}c@{}}100.0\% $-$\\ 1.564e5 $-$\end{tabular}
& \begin{tabular}[c]{@{}c@{}}76.44\% $-$\\ \textbf{2.000e5} $\approx$\end{tabular}
& \begin{tabular}[c]{@{}c@{}}95.39\% $-$\\ \textbf{2.000e5} $\approx$\end{tabular}
& \begin{tabular}[c]{@{}c@{}}\textbf{100.0}\% $\approx$\\ {2.000e5} $\approx$\end{tabular}
& \begin{tabular}[c]{@{}c@{}}\textbf{100.0}\% $\approx$\\ 2.000e5 $-$\end{tabular}
& \begin{tabular}[c]{@{}c@{}}98.91\% $-$\\ \textbf{2.000e5} $\approx$\end{tabular}
& \begin{tabular}[c]{@{}c@{}}\textbf{100.0}\% $\approx$\\ \textbf{2.000e5} $\approx$\end{tabular}
& \begin{tabular}[c]{@{}c@{}}77.23\% $-$\\ 1.861e5 $-$\end{tabular}
& \begin{tabular}[c]{@{}c@{}}83.74\% $-$\\ 1.895e5 $-$\end{tabular}
& \begin{tabular}[c]{@{}c@{}}88.28\% $-$\\ 1.933e5 $-$\end{tabular}
& \begin{tabular}[c]{@{}c@{}}92.20\% $-$\\ 1.960e5 $-$\end{tabular}
& 9 / 2 / 0
\\ 
\hline
\end{tabular}
}
\label{tb:r+s}
\end{table*}

\begin{table*}[]
\centering
\caption{
Comparing RL-DAS with comparison algorithms on Shifted \& Rotated 20D problems.}
\resizebox{0.98\textwidth}{!}{%
\begin{tabular}{cc|c|lllllllllll|c}
\hline
  \multicolumn{2}{c|}{\multirow{3}{*}{Taxonomy}} 
& \multicolumn{1}{c|}{\multirow{3}{*}{Algorithm}} 
& \multicolumn{11}{c|}{\multirow{1}{*}{Problem Class}}
& \multicolumn{1}{c}{\multirow{3}{*}{\begin{tabular}[c]{@{}c@{}}Significance\\Test\\(Win/Tie/Loss)\end{tabular}}}
\\ \cline{4-14}
 & &
 & \multicolumn{1}{l}{\multirow{2}{*}{\quad C1}}
 & \multicolumn{1}{l}{\multirow{2}{*}{\quad C2}}
 & \multicolumn{1}{l}{\multirow{2}{*}{\quad C3}}
 & \multicolumn{1}{l}{\multirow{2}{*}{\quad C4}}
 & \multicolumn{1}{l}{\multirow{2}{*}{\quad C5}}
 & \multicolumn{1}{l}{\multirow{2}{*}{\quad C6}}
 & \multicolumn{1}{l}{\multirow{2}{*}{\quad C7}}
 & \multicolumn{1}{l}{\multirow{2}{*}{\quad C8}}
 & \multicolumn{1}{l}{\multirow{2}{*}{\quad C9}}
 & \multicolumn{1}{l}{\multirow{2}{*}{\quad C10}}
 & \multicolumn{1}{l|}{\multirow{2}{*}{\quad C11}}
 & 
 \\
 & & & & & & & & & & & & & &
 \\ \hline
\multicolumn{2}{c|}{\multirow{3}{*}{DAS}}
& \textbf{RL-DAS}
& \begin{tabular}[c]{@{}c@{}}\textbf{100.0}\%\\ 3.547e5\end{tabular}
& \begin{tabular}[c]{@{}c@{}}\textbf{76.54}\%\\ \textbf{1.000e6}\end{tabular}
& \begin{tabular}[c]{@{}c@{}}\textbf{96.89}\%\\ \textbf{1.000e6}\end{tabular}
& \begin{tabular}[c]{@{}c@{}}\textbf{100.0}\%\\ \textbf{1.000e6}\end{tabular}
& \begin{tabular}[c]{@{}c@{}}\textbf{100.0}\%\\ \textbf{1.000e6}\end{tabular}
& \begin{tabular}[c]{@{}c@{}}\textbf{97.91}\%\\ \textbf{1.000e6}\end{tabular}
& \begin{tabular}[c]{@{}c@{}}\textbf{100.0}\%\\ \textbf{1.000e6}\end{tabular}
& \begin{tabular}[c]{@{}c@{}}\textbf{79.59}\%\\ \textbf{8.489e5}\end{tabular}
& \begin{tabular}[c]{@{}c@{}}\textbf{87.55}\%\\ \textbf{9.211e5}\end{tabular}
& \begin{tabular}[c]{@{}c@{}}89.74\%\\ \textbf{9.112e5}\end{tabular}
& \begin{tabular}[c]{@{}c@{}}\textbf{92.87}\%\\ 9.231e5\end{tabular}
& - / - / -
\\
&
& \textbf{Rand-DAS}
& \begin{tabular}[c]{@{}c@{}}\textbf{100.0}\% $\approx$\\ 6.856e5 $-$\end{tabular}
& \begin{tabular}[c]{@{}c@{}}73.70\% $-$\\ \textbf{1.000e6} $\approx$\end{tabular}
& \begin{tabular}[c]{@{}c@{}}96.83\% $\approx$\\ \textbf{1.000e6} $\approx$\end{tabular}
& \begin{tabular}[c]{@{}c@{}}\textbf{100.0}\% $\approx$\\ \textbf{1.000e6} $\approx$\end{tabular}
& \begin{tabular}[c]{@{}c@{}}\textbf{100.0}\% $\approx$\\ \textbf{1.000e6} $\approx$\end{tabular}
& \begin{tabular}[c]{@{}c@{}}96.51\% $-$\\ \textbf{1.000e6} $\approx$\end{tabular}
& \begin{tabular}[c]{@{}c@{}}\textbf{100.0}\% $\approx$\\ \textbf{1.000e6} $\approx$\end{tabular}
& \begin{tabular}[c]{@{}c@{}}79.14\% $-$\\ \textbf{9.653e5} $+$\end{tabular}
& \begin{tabular}[c]{@{}c@{}}87.03\% $-$\\ \textbf{9.456e5} $\approx$\end{tabular}
& \begin{tabular}[c]{@{}c@{}}89.48\% $-$\\ \textbf{9.520e5} $\approx$\end{tabular}
& \begin{tabular}[c]{@{}c@{}}92.35\% $-$\\ 9.347e5 $-$\end{tabular}
& 7 / 4 / 0
\\ \hline
\multicolumn{2}{c|}{\multirow{1}{*}{AS}}
& \textbf{AS$^*$}
& \begin{tabular}[c]{@{}c@{}}\textbf{100.0}\% $\approx$\\3.635e5 $-$\end{tabular}
& \begin{tabular}[c]{@{}c@{}}75.85\% $-$\\ \textbf{1.000e6} $\approx$\end{tabular}
& \begin{tabular}[c]{@{}c@{}}96.49\% $-$\\ \textbf{1.000e6} $\approx$\end{tabular}
& \begin{tabular}[c]{@{}c@{}}\textbf{100.0}\% $\approx$\\ 1.000e6 $\approx$\end{tabular}
& \begin{tabular}[c]{@{}c@{}}\textbf{100.0}\% $\approx$\\ \textbf{1.000e6} $\approx$\end{tabular}
& \begin{tabular}[c]{@{}c@{}}97.62\% $\approx$\\ \textbf{1.000e6} $\approx$\end{tabular}
& \begin{tabular}[c]{@{}c@{}}\textbf{100.0}\% $\approx$\\ \textbf{1.000e6} $\approx$\end{tabular}
& \begin{tabular}[c]{@{}c@{}}78.86\% $-$\\ 8.763e5 $-$\end{tabular}
& \begin{tabular}[c]{@{}c@{}}87.13\% $-$\\ 9.137e5 $\approx$\end{tabular}
& \begin{tabular}[c]{@{}c@{}}89.39\% $-$\\ 9.276e5 $\approx$\end{tabular}
& \begin{tabular}[c]{@{}c@{}}92.57\% $-$\\ \textbf{8.973e5} $+$\end{tabular}
& 8 / 3 / 0
\\ \hline
\multicolumn{2}{c|}{\multirow{5}{*}{Backbones}}
& \textbf{JDE21}
& \begin{tabular}[c]{@{}c@{}}\textbf{100.0}\% $\approx$\\ 9.573e5 $-$\end{tabular}
& \begin{tabular}[c]{@{}c@{}}62.73\% $-$\\ \textbf{1.000e6} $\approx$\end{tabular}
& \begin{tabular}[c]{@{}c@{}}96.24\% $-$\\ \textbf{1.000e6} $\approx$\end{tabular}
& \begin{tabular}[c]{@{}c@{}}\textbf{100.0}\% $\approx$\\ \textbf{1.000e6} $\approx$\end{tabular}
& \begin{tabular}[c]{@{}c@{}}\textbf{100.0}\% $\approx$\\ \textbf{1.000e6} $\approx$\end{tabular}
& \begin{tabular}[c]{@{}c@{}}95.05\% $-$\\ \textbf{1.000e6} $\approx$\end{tabular}
& \begin{tabular}[c]{@{}c@{}}\textbf{100.0}\% $\approx$\\ \textbf{1.000e6} $\approx$\end{tabular}
& \begin{tabular}[c]{@{}c@{}}77.72\% $-$\\ 9.619e5 $-$\end{tabular}
& \begin{tabular}[c]{@{}c@{}}86.55\% $-$\\ \textbf{9.432e5} $\approx$\end{tabular}
& \begin{tabular}[c]{@{}c@{}}89.32\% $-$\\ \textbf{9.241e5} $\approx$\end{tabular}
& \begin{tabular}[c]{@{}c@{}}90.86\% $-$\\ 9.880e5 $-$\end{tabular}
& 8 / 3 / 0
\\
&
& \textbf{MadDE}
& \begin{tabular}[c]{@{}c@{}}\textbf{100.0}\% $\approx$\\ 3.681e5 $-$\end{tabular}
& \begin{tabular}[c]{@{}c@{}}75.76\% $-$\\ \textbf{1.000e6} $\approx$\end{tabular}
& \begin{tabular}[c]{@{}c@{}}96.53\% $-$\\ \textbf{1.000e6} $\approx$\end{tabular}
& \begin{tabular}[c]{@{}c@{}}\textbf{100.0}\% $\approx$\\ \textbf{1.000e6} $\approx$\end{tabular}
& \begin{tabular}[c]{@{}c@{}}\textbf{100.0}\% $\approx$\\ \textbf{1.000e6} $\approx$\end{tabular}
& \begin{tabular}[c]{@{}c@{}}97.62\% $\approx$\\ \textbf{1.000e6} $\approx$\end{tabular}
& \begin{tabular}[c]{@{}c@{}}\textbf{100.0}\% $\approx$\\ \textbf{1.000e6} $\approx$\end{tabular}
& \begin{tabular}[c]{@{}c@{}}78.21\% $-$\\ 8.787e5 $-$\end{tabular}
& \begin{tabular}[c]{@{}c@{}}86.66\% $-$\\ \textbf{9.202e5} $\approx$\end{tabular}
& \begin{tabular}[c]{@{}c@{}}89.03\% $-$\\ \textbf{9.486e5} $\approx$\end{tabular}
& \begin{tabular}[c]{@{}c@{}}92.64\% $-$\\ 9.022e5 $+$\end{tabular}
& 8 / 3 / 0
\\
&
& \textbf{NL-SHADE-RSP}
& \begin{tabular}[c]{@{}c@{}}\textbf{100.0}\% $\approx$\\ {1.000e6} $-$\end{tabular}
& \begin{tabular}[c]{@{}c@{}}70.57\% $-$\\ \textbf{1.000e6} $\approx$\end{tabular}
& \begin{tabular}[c]{@{}c@{}}95.57\% $-$\\ \textbf{1.000e6} $\approx$\end{tabular}
& \begin{tabular}[c]{@{}c@{}}\textbf{100.0}\% $\approx$\\ \textbf{1.000e6} $\approx$\end{tabular}
& \begin{tabular}[c]{@{}c@{}}\textbf{100.0}\% $\approx$\\ \textbf{1.000e6} $\approx$\end{tabular}
& \begin{tabular}[c]{@{}c@{}}92.94\% $-$\\ \textbf{1.000e6} $\approx$\end{tabular}
& \begin{tabular}[c]{@{}c@{}}\textbf{100.0}\% $\approx$\\ \textbf{1.000e6} $\approx$\end{tabular}
& \begin{tabular}[c]{@{}c@{}}77.13\% $-$\\ 9.653e5 $-$\end{tabular}
& \begin{tabular}[c]{@{}c@{}}86.35\% $-$\\ \textbf{9.505e5} $\approx$\end{tabular}
& \begin{tabular}[c]{@{}c@{}}87.36\% $-$\\ \textbf{9.641e5} $\approx$\end{tabular}
& \begin{tabular}[c]{@{}c@{}}91.11\% $-$\\ 9.923e5 $-$\end{tabular}
& 8 / 3 / 0
\\ \hline
\multicolumn{1}{c}{\multirow{9}{*}{\begin{tabular}[c]{@{}c@{}}Other\\Advanced\\DE\\Variants\end{tabular}}}
& \multicolumn{1}{|c|}{\multirow{1}{*}{Adaptive}}
& \textbf{EPSDE}
& \begin{tabular}[c]{@{}c@{}}\textbf{100.0}\% $\approx$\\ 4.724e5 $-$\end{tabular}
& \begin{tabular}[c]{@{}c@{}}64.31\% $-$\\ \textbf{1.000e6} $\approx$\end{tabular}
& \begin{tabular}[c]{@{}c@{}}96.38\% $-$\\ \textbf{1.000e6} $\approx$\end{tabular}
& \begin{tabular}[c]{@{}c@{}}\textbf{100.0}\% $\approx$\\ \textbf{1.000e6} $\approx$\end{tabular}
& \begin{tabular}[c]{@{}c@{}}\textbf{100.0}\% $\approx$\\ \textbf{1.000e6} $\approx$\end{tabular}
& \begin{tabular}[c]{@{}c@{}}97.24\% $-$\\ \textbf{1.000e6} $\approx$\end{tabular}
& \begin{tabular}[c]{@{}c@{}}\textbf{100.0}\% $\approx$\\ \textbf{1.000e6} $\approx$\end{tabular}
& \begin{tabular}[c]{@{}c@{}}78.83\% $-$\\ 9.588e5 $-$\end{tabular}
& \begin{tabular}[c]{@{}c@{}}86.45\% $-$\\ \textbf{9.462e5} $\approx$\end{tabular}
& \begin{tabular}[c]{@{}c@{}}89.95\% $+$\\ \textbf{9.031e5} $\approx$\end{tabular}
& \begin{tabular}[c]{@{}c@{}}91.52\% $-$\\ 9.471e5 $-$\end{tabular}
& 7 / 3 / 1
\\ \cline{2-15}
& \multicolumn{1}{|c|}{\multirow{1}{*}{Ensemble}}
& \textbf{EDEV}
& \begin{tabular}[c]{@{}c@{}}\textbf{100.0}\% $\approx$\\ 4.545e5 $-$\end{tabular}
& \begin{tabular}[c]{@{}c@{}}66.16\% $-$\\ \textbf{1.000e6} $\approx$\end{tabular}
& \begin{tabular}[c]{@{}c@{}}96.43\% $-$\\ \textbf{1.000e6} $\approx$\end{tabular}
& \begin{tabular}[c]{@{}c@{}}\textbf{100.0}\% $\approx$\\ \textbf{1.000e6} $\approx$\end{tabular}
& \begin{tabular}[c]{@{}c@{}}\textbf{100.0}\% $\approx$\\ \textbf{1.000e6} $\approx$\end{tabular}
& \begin{tabular}[c]{@{}c@{}}97.32\% $-$\\ \textbf{1.000e6} $\approx$\end{tabular}
& \begin{tabular}[c]{@{}c@{}}\textbf{100.0}\% $\approx$\\ \textbf{1.000e6} $\approx$\end{tabular}
& \begin{tabular}[c]{@{}c@{}}78.73\% $-$\\ 9.499e5 $-$\end{tabular}
& \begin{tabular}[c]{@{}c@{}}86.56\% $-$\\ \textbf{9.351e5} $\approx$\end{tabular}
& \begin{tabular}[c]{@{}c@{}}\textbf{89.97}\% $+$\\ \textbf{8.989e5} $\approx$\end{tabular}
& \begin{tabular}[c]{@{}c@{}}91.34\% $-$\\ 9.414e5 $-$\end{tabular}
& 7 / 3 / 1
\\ \cline{2-15}
& \multicolumn{1}{|c|}{\multirow{5}{*}{\begin{tabular}[c]{@{}c@{}}RL\\Assisted\end{tabular}}}
& \textbf{DEDQN}
& \begin{tabular}[c]{@{}c@{}}100.0\% $-$\\ 9.945e5 $-$\end{tabular}
& \begin{tabular}[c]{@{}c@{}}63.74\% $-$\\ \textbf{1.000e6} $\approx$\end{tabular}
& \begin{tabular}[c]{@{}c@{}}95.38\% $-$\\ \textbf{1.000e6} $\approx$\end{tabular}
& \begin{tabular}[c]{@{}c@{}}\textbf{100.0}\% $\approx$\\ \textbf{1.000e6} $\approx$\end{tabular}
& \begin{tabular}[c]{@{}c@{}}\textbf{100.0}\% $\approx$\\ \textbf{1.000e6} $\approx$\end{tabular}
& \begin{tabular}[c]{@{}c@{}}96.11\% $-$\\ \textbf{1.000e6} $\approx$\end{tabular}
& \begin{tabular}[c]{@{}c@{}}\textbf{100.0}\% $\approx$\\ \textbf{1.000e6} $\approx$\end{tabular}
& \begin{tabular}[c]{@{}c@{}}75.15\% $-$\\ 9.812e5 $-$\end{tabular}
& \begin{tabular}[c]{@{}c@{}}85.53\% $-$\\ \textbf{9.687e5} $\approx$\end{tabular}
& \begin{tabular}[c]{@{}c@{}}88.01\% $-$\\ \textbf{9.544e5} $\approx$\end{tabular}
& \begin{tabular}[c]{@{}c@{}}90.90\% $-$\\ 9.962e5 $-$\end{tabular}
& 8 / 3 / 0
\\
& \multicolumn{1}{|c|}{}
& \textbf{DEDDQN}
& \begin{tabular}[c]{@{}c@{}}100.0\% $-$\\ 9.813e5 $-$\end{tabular}
& \begin{tabular}[c]{@{}c@{}}62.19\% $-$\\ \textbf{1.000e6} $\approx$\end{tabular}
& \begin{tabular}[c]{@{}c@{}}95.15\% $-$\\ \textbf{1.000e6} $\approx$\end{tabular}
& \begin{tabular}[c]{@{}c@{}}\textbf{100.0}\% $\approx$\\ \textbf{1.000e6} $\approx$\end{tabular}
& \begin{tabular}[c]{@{}c@{}}\textbf{100.0}\% $\approx$\\ \textbf{1.000e6} $\approx$\end{tabular}
& \begin{tabular}[c]{@{}c@{}}96.54\% $-$\\ \textbf{1.000e6} $\approx$\end{tabular}
& \begin{tabular}[c]{@{}c@{}}\textbf{100.0}\% $\approx$\\ \textbf{1.000e6} $\approx$\end{tabular}
& \begin{tabular}[c]{@{}c@{}}75.91\% $-$\\ 9.735e5 $-$\end{tabular}
& \begin{tabular}[c]{@{}c@{}}86.03\% $-$\\ \textbf{9.625e5} $\approx$\end{tabular}
& \begin{tabular}[c]{@{}c@{}}88.01\% $-$\\ \textbf{9.519e5} $\approx$\end{tabular}
& \begin{tabular}[c]{@{}c@{}}91.05\% $-$\\ 9.950e5 $-$\end{tabular}
& 8 / 3 / 0
\\
& \multicolumn{1}{|c|}{}
& \textbf{LDE}
& \begin{tabular}[c]{@{}c@{}}\textbf{100.0}\% $\approx$\\ 5.642e5 $-$\end{tabular}
& \begin{tabular}[c]{@{}c@{}}61.22\% $-$\\ \textbf{1.000e6} $\approx$\end{tabular}
& \begin{tabular}[c]{@{}c@{}}95.96\% $-$\\ \textbf{1.000e6} $\approx$\end{tabular}
& \begin{tabular}[c]{@{}c@{}}\textbf{100.0}\% $\approx$\\ \textbf{1.000e6} $\approx$\end{tabular}
& \begin{tabular}[c]{@{}c@{}}\textbf{100.0}\% $\approx$\\ \textbf{1.000e6} $\approx$\end{tabular}
& \begin{tabular}[c]{@{}c@{}}97.54\% $-$\\ \textbf{1.000e6} $\approx$\end{tabular}
& \begin{tabular}[c]{@{}c@{}}\textbf{100.0}\% $\approx$\\ \textbf{1.000e6} $\approx$\end{tabular}
& \begin{tabular}[c]{@{}c@{}}77.15\% $-$\\ 9.287e5 $-$\end{tabular}
& \begin{tabular}[c]{@{}c@{}}86.30\% $-$\\ \textbf{9.530e5} $\approx$\end{tabular}
& \begin{tabular}[c]{@{}c@{}}88.52\% $-$\\ \textbf{9.433e5} $\approx$\end{tabular}
& \begin{tabular}[c]{@{}c@{}}91.33\% $-$\\ 9.487e5 $-$\end{tabular}
& 8 / 3 / 0
\\ 
\hline
\end{tabular}
}
\label{tb:20d}
\end{table*}

\subsubsection{Comparison with Backbones}

Table~\ref{tb:r+s} and Table~\ref{tb:20d} show the performance of algorithms on 10D and 20D problems respectively, 
where the upper value is the descent which describes the average descent of costs in percentage (normalized by the best costs in the initial populations, which are consistent among algorithms with random number seeds), 
and the lower value is FEs describes the average consumed function evaluations to reach the termination criterion. The `$+$', `$-$' and `$\approx$' indicate the statistical test results that the performance of competing methods is better, worse and no significant difference with RL-DAS respectively according to the Wilcoxon rank-sum test at the 0.05 significance level. The last columns of the two tables summarize the statistical results where the first value counts the cases when RL-DAS significantly outperforms the competitor (the two tests contain at least one `$-$' while the test on descent is not `$+$'); the second value counts the cases that there is no significant difference between RL-DAS and the competitor (both tests are `$\approx$'); and the third value counts the cases otherwise. From the results in these tables, we can conclude that:
\begin{itemize}
    \item RL-DAS surpasses the Rand-DAS and defeats the backbones on most problems in both 10D and 20D scenarios, which verifies the effectiveness of RL-DAS. Note that the experiment presents the average performance over the instance sets without assuming that RL-DAS defeats the comparison algorithms in all cases.
    \item For the 10D scenario, while the previously reported results~\cite{cec2021TR} show that the JDE21, NL-SHADE-RSP and MadDE algorithms can optimize the original benchmark problems towards very low costs, their performance on the augmented benchmarks is far from this. Our results indicate that the algorithms may overfit to the small group of problems provided in the original benchmark, resulting in degraded performance when applied to other shifting and rotation configurations. 
    These complex problems require a better balance between exploration and exploitation to avoid falling into local optima. The three backbone algorithms work with their manual exploration-exploitation balance methods, which give them unique searching characteristics and make them winners of the CEC2021 competition. However, these manual designs are based on a small group of benchmark problems and fail to show enough adaptiveness in more test instances. On the contrary, RL-DAS employs an DRL agent to dynamically switch these algorithms to achieve comprehensively better performance.
    \item For the 20D problems shown in Table~\ref{tb:20d}, RL-DAS still outperforms the others. The increased problem space makes the optimization more difficult, and our relative performance over the competitors is even better than that on 10D problems. The results indicate that dimensions do not affect the RL-DAS much, the conclusions on 10D problems are still applicable to the 20D benchmark.
    \item Furthermore, an intriguing observation is the improved performance of the Rand-DAS over the three backbone algorithms. 
    This further validates the importance of DAS over traditional AS setting: a single hand-crafted algorithm can hardly be efficient in solving all problem cases (as indicated by the NFL theorem), and even 
    a random combination of a few algorithms has the potential to 
    achieve complementary performance. Nevertheless, the comparison between RL-DAS and Rand-DAS shows that our RL agent further exploits this potential and shows significant improvement over the random agent.
\end{itemize}

\subsubsection{Comparison between DAS and AS}

The DAS and backbones parts of Table~\ref{tb:r+s} and Table~\ref{tb:20d} verify that RL-DAS outperforms the compared single algorithm. Then a question comes out: can RL-DAS defeat other methods that combine these single algorithms, for example, by the Algorithm Selection (AS)? To figure out the answer, we introduce a competitor that takes the best result among the three backbone algorithms as its performance, namely AS$^*$, to simulate the performance upper bound of AS methods. For each problem instance $i$, the result of AS$^*$ is calculated by:
\begin{equation}
    \text{AS}^*(i) = \max(\text{JDE21}(i), \text{MadDE}(i), \text{NL-SHADE-RSP}(i))
\end{equation}
where JDE21($i$) indicates the optimization result of JDE21 on instance $i$, and so do the items for MadDE and NL-SHADE-RSP. The performance comparison is shown in the AS parts of Table~\ref{tb:r+s} and Table~\ref{tb:20d}.

It can be seen that RL-DAS outperforms AS$^*$ in most of the problems. AS methods employ a single candidate algorithm for each problem instance, therefore the upper bound of its performance can not exceed the best performance among all candidate algorithms, let alone that practical AS methods usually require external evaluation resources to select an algorithm~\cite{kerschke2019automated} which will inevitably impact the performance. 
Differently, the DAS methods select an algorithm at each interval which has an opportunity to present a complementary performance and breaks through the single best candidate algorithm. The outstanding performance of Rand-DAS in Table~\ref{tb:r+s} and Table~\ref{tb:20d} prove its potential. The employment of an DRL agent ensures wise selection thus RL-DAS defeats AS$^*$ significantly.

\subsubsection{Comparison with Other Advanced DE Variants}

The last part in Table~\ref{tb:r+s} and Table~\ref{tb:20d} present the comparison between our RL-DAS and other advanced DE. For the sake of fairness, the three RL-assisted methods are trained on the same training set as RL-DAS and update their models by equal steps.

RL-DAS demonstrates a significant advantage over adaptive and ensemble algorithms for both 10D and 20D problems. Additionally, for the three methods utilizing RL agents, namely DEDQN, DEDDQN, and LDE, RL-DAS also outperforms them. This can be attributed to RL-DAS's integration of three advanced algorithms, which incorporates the existing expert knowledge for a promising performance. 
\begin{table}[]
\centering
\caption{Computational cost comparison.}
\resizebox{0.4\columnwidth}{!}{%
\begin{tabular}{ccccccc}
\toprule
\textbf{Algorithm}
& \textbf{JDE21}
& \textbf{MadDE}
& \textbf{NL-SHADE-RSP}
& \textbf{EPSDE}
& \textbf{EDEV}
\\ \midrule
\textbf{Time (s)}
& 0.748
& 0.905
& 0.791
& 0.534
& 0.717
\\ \midrule
\textbf{Algorithm}
& \textbf{RL-DAS}
& \textbf{DEDQN}
& \textbf{DEDDQN}
& \textbf{LDE}
\\ \midrule
\textbf{Time (s)}
& 1.434
& 0.986
& 66.891
& 1.263
\\ \bottomrule
\end{tabular}%
}
\label{tb:timecost}
\end{table}
\subsubsection{Runtime Comparison}
Table~\ref{tb:timecost} presents the average computational time of different algorithms on 10D C2 (Schwefel) testing set problems. The upper row represents traditional non-learning algorithms, while the lower row presents the RL-assisted algorithms including our RL-DAS. It is evident that the computational costs of most DE algorithms are comparable, with MadDE being relatively slower due to its use of more operators and parameter adaptive mechanisms. RL-assisted algorithms, in general, require more time for network inference. Among them, DEDDQN takes the longest time, attributed to its costly and complex 99-dimensional state representation and one-by-one operator selection for each individual. RL-DAS, on the other hand, completes optimization in 1.434 seconds for 200,000 FEs. While it requires more time for network inference and algorithm switching compared to JDE21, MadDE, and NL-SHADE-RSP, the additional computational cost is generally acceptable regarding the state-of-the-art optimization accuracy of RL-DAS.

In summary, RL-DAS achieves state-of-the-art optimization performance, it outperforms not only the backbones but also the advanced DE variants including adaptive DE and RL-assisted DE, while incurring an additional acceptable computational cost. To unveil the secret behind this performance, we conduct an in-depth analysis of the learned policy, as detailed in Section 3 of our supplementary document.

\begin{table}[t]
\centering
\caption{Transfer results from one class of problems to others.}
\resizebox{0.6\columnwidth}{!}{%
\begin{tabular}{ccccccccc}
\hline
\multicolumn{1}{c|}{\begin{tabular}[c]{@{}c@{}}Problem\\Class\end{tabular}}
& \multicolumn{2}{c|}{\begin{tabular}[c]{@{}c@{}}\textbf{RL-DAS}\\\textbf{Zero-shot (C1)}\end{tabular}}
& \multicolumn{2}{c|}{\begin{tabular}[c]{@{}c@{}}\textbf{RL-DAS}\\\textbf{Zero-shot (C2)}\end{tabular}}

& \multicolumn{2}{c|}{\begin{tabular}[c]{@{}c@{}}\textbf{RL-DAS}\\\textbf{Zero-shot (C11)}\end{tabular}}

& \multicolumn{2}{c}{\begin{tabular}[c]{@{}c@{}}\textbf{RL-DAS}\end{tabular}}

\\ \hline
\multicolumn{1}{c|}{C1}
& $\backslash$             & \multicolumn{1}{c|}{$\backslash$}
& \textbf{100.0}\% $\approx$             & \multicolumn{1}{c|}{\textbf{1.046e5} $+$}

& \textbf{100.0}\% $\approx$   & \multicolumn{1}{c|}{1.215e5 $-$}
& \textbf{100.0}\%             & \multicolumn{1}{c}{1.127e5}

\\
\multicolumn{1}{c|}{C2}
& 84.32\% $-$             & \multicolumn{1}{c|}{\textbf{2.000e5}  $\approx$}
& $\backslash$             & \multicolumn{1}{c|}{$\backslash$}

& 83.69\% $-$             & \multicolumn{1}{c|}{\textbf{2.000e5} $\approx$}
& \textbf{85.97}\%             & \multicolumn{1}{c}{\textbf{2.000e5}}

\\
\multicolumn{1}{c|}{C3}
& 95.67\% $\approx$                      & \multicolumn{1}{c|}{\textbf{2.000e5}  $\approx$}
& 95.61\% $\approx$             & \multicolumn{1}{c|}{\textbf{2.000e5} $\approx$}

& \textbf{95.86}\% $+$             & \multicolumn{1}{c|}{\textbf{2.000e5} $\approx$}
& 95.63\%             & \multicolumn{1}{c}{\textbf{2.000e5}}

\\
\multicolumn{1}{c|}{C4}
& \textbf{100.0}\% $\approx$             & \multicolumn{1}{c|}{1.999e5 $\approx$}
& \textbf{100.0}\% $\approx$             & \multicolumn{1}{c|}{\textbf{1.996e5} $+$}

& \textbf{100.0}\% $\approx$   & \multicolumn{1}{c|}{2.000e5 $\approx$}
& \textbf{100.0}\%             & \multicolumn{1}{c}{2.000e5}

\\
\multicolumn{1}{c|}{C5}
& \textbf{100.0}\% $\approx$             & \multicolumn{1}{c|}{1.996e5 $\approx$}
& \textbf{100.0}\% $\approx$             & \multicolumn{1}{c|}{\textbf{1.986e5} $+$}

& \textbf{100.0}\% $\approx$   & \multicolumn{1}{c|}{1.997e5 $\approx$}
& \textbf{100.0}\%             & \multicolumn{1}{c}{1.996e5}

\\
\multicolumn{1}{c|}{C6}
& 99.37\% $\approx$                      & \multicolumn{1}{c|}{\textbf{2.000e5} $\approx$}
& \textbf{99.56}\% $+$             & \multicolumn{1}{c|}{\textbf{2.000e5} $\approx$}

& 99.38\% $\approx$               & \multicolumn{1}{c|}{\textbf{2.000e5} $\approx$}
& 99.39\%                      & \multicolumn{1}{c}{\textbf{2.000e5}}

\\
\multicolumn{1}{c|}{C7}
& \textbf{100.0}\% $\approx$             & \multicolumn{1}{c|}{\textbf{2.000e5} $\approx$}
& \textbf{100.0}\% $\approx$             & \multicolumn{1}{c|}{\textbf{2.000e5} $\approx$}

& \textbf{100.0}\% $\approx$   & \multicolumn{1}{c|}{\textbf{2.000e5} $\approx$}
& \textbf{100.0}\%             & \multicolumn{1}{c}{\textbf{2.000e5}}

\\
\multicolumn{1}{c|}{C8}
& \textbf{80.63}\% $+$                      & \multicolumn{1}{c|}{1.801e5 $+$}
& 80.27\% $-$                      & \multicolumn{1}{c|}{1.811e5 $+$}

& 80.29\% $-$        & \multicolumn{1}{c|}{1.830e5 $-$}
& 80.48\%             & \multicolumn{1}{c}{1.821e5}

\\
\multicolumn{1}{c|}{C9}
& 90.14\% $\approx$                      & \multicolumn{1}{c|}{1.868e5 $+$}
& 89.73\% $-$                      & \multicolumn{1}{c|}{\textbf{1.873e5} $+$}

& 89.93\% $-$             & \multicolumn{1}{c|}{1.889e5 $-$}
& \textbf{90.18}\%            & \multicolumn{1}{c}{1.882e5}

\\
\multicolumn{1}{c|}{C10}
& 88.32\% $-$                      & \multicolumn{1}{c|}{\textbf{1.909e5} $+$}
& \textbf{88.81}\% $\approx$             & \multicolumn{1}{c|}{\textbf{1.909e5} $+$}

& 88.59\% $-$             & \multicolumn{1}{c|}{1.921e5 $-$}
& \textbf{88.79}\%            & \multicolumn{1}{c}{1.914e5}

\\ \hline
\multicolumn{1}{c|}{}
& \multicolumn{2}{c|}{2 / 5 / 2}
& \multicolumn{2}{c|}{2 / 2 / 5}
& \multicolumn{2}{c|}{5 / 4 / 1}
& \multicolumn{2}{c}{- / - / -}

\\
\end{tabular}%
}
\label{tb:zeroshot}
\vspace{-2mm}
\end{table}

\begin{table}[t]
    \centering
    \caption{Transfer results under different problem class partitions.}
    \resizebox{0.6\columnwidth}{!}{%
    \begin{tabular}{ccccccccc}
    \hline
    \multicolumn{1}{c|}{\begin{tabular}[c]{@{}c@{}}Transfer\\Setting\end{tabular}}
    & \multicolumn{2}{c|}{\begin{tabular}[c]{@{}c@{}}\textbf{NL-SHADE-RSP}\end{tabular}}
    & \multicolumn{2}{c|}{\begin{tabular}[c]{@{}c@{}}\textbf{MadDE}\end{tabular}}
    & \multicolumn{2}{c|}{\begin{tabular}[c]{@{}c@{}}\textbf{JDE21}\end{tabular}}
    & \multicolumn{2}{c}{\begin{tabular}[c]{@{}c@{}}\textbf{RL-DAS}\\\textbf{(Zero-shot)}\end{tabular}}
    
    \\ \hline
    \multicolumn{1}{c|}{TS1}
    & 88.38\% $-$             & \multicolumn{1}{c|}{2.000e5 $-$}
    & 92.62\% $-$             & \multicolumn{1}{c|}{1.999e5 $-$}
    & 89.87\% $-$             & \multicolumn{1}{c|}{2.000e5 $-$}
    & \textbf{92.86}\%                       & \multicolumn{1}{c}{\textbf{1.991e5}}
    
    \\
    \multicolumn{1}{c|}{TS2}
    & 84.88\% $-$                           & \multicolumn{1}{c|}{2.000e5 $-$}
    & 89.47\% $-$                           & \multicolumn{1}{c|}{\textbf{1.961e5} $+$}
    & 86.12\% $-$                           & \multicolumn{1}{c|}{1.985e5 $-$}
    & \textbf{89.62}\%                      & \multicolumn{1}{c}{1.969e5}
    
    \\
    \multicolumn{1}{c|}{TS3}
    & 90.76\% $-$                     & \multicolumn{1}{c|}{2.000e5 $-$}
    & 91.47\% $-$                     & \multicolumn{1}{c|}{1.951e5 $-$}
    & 90.86\% $-$                      & \multicolumn{1}{c|}{1.988e5 $-$}
    & \textbf{91.65}\%                 & \multicolumn{1}{c}{\textbf{1.917e5}}
    
    \\
    
    
    \hline
    \multicolumn{1}{c|}{}
    & \multicolumn{2}{c|}{3 / 0 / 0}
    & \multicolumn{2}{c|}{3 / 0 / 0}
    & \multicolumn{2}{c|}{3 / 0 / 0}
    & \multicolumn{2}{c}{- / - / -}
    \\
    \end{tabular}%
    }
    \label{tb:noisy}
    \end{table}
\subsection{Zero-shot Knowledge Transfer}\label{Exp:ZT}

In this section, we conduct two types of zero-shot knowledge transfer experiments. First, we choose models from the three problem classes which have significant performance differences as reported in Section~\ref{Exp:PC}: the unimodal problem C1 (Bent Cigar), the basic multimodal problem C2 (Schwefel) and the mix-class problems C11, and then zero-shot transfer them to the other CEC problems. The results are shown in Table~\ref{tb:zeroshot}.
It can be seen that RL-DAS successfully transfers its learned knowledge to unseen problems, which allows the models from C1, C2 and C11 achieve close results to the original RL-DAS. 

In the second experiment, we establish different transfer settings, denoted as TS1-TS3, by partitioning the ten classes of problems into training and testing sets with varying proportions: 2 : 8~(F1-F2 for training, F3-F10 for testing), 5 : 5~(F1-F5 for training, F6-F10 for testing), and 8 : 2~(F1-F8 for training, F9-F10 for testing), respectively. 
The training and testing sets, with instances mixed from different problem classes, are generated in the same way as C11. The results are shown in Table~\ref{tb:noisy}, where the three backbone algorithms serve as references.
RL-DAS outperforms the baselines on most of transfer settings in Table~\ref{tb:noisy}. It demonstrates promising generalization ability, transferring the knowledge learned from a small training problem set to a much larger and more complex problem set.

The results in the two tables collectively show that the RL agent captures the intrinsic characteristics of problems and the searching strategies of the backbone algorithms. This enables the agent to generalize its knowledge to unseen problems and achieve state-of-the-art performance.

\subsection{Ablation Study and Hyperparameter Research}\label{Exp:AS}
\subsubsection{Ablation Study}

In order to verify the necessity of the framework components, 
we conducted ablation studies on the state features and the context memory. For the features, we removed the embedding and concatenation of the LA feature (named RL-DAS w/o LA) and the AH feature (named RL-DAS w/o AH), respectively, and tested them on 10D problems. The ablation study that removes both features is equivalent to the `Rand-DAS' condition studied in previous experiments, as the model cannot function without input. 
Besides, we ablate the context memory (named RL-DAS w/o Context) to investigate the effect of the warm start procedure. In this scenario, the parameter memories, elite archives and other adaption mechanisms are cleaned and re-initialized at the beginning of each internal.

Table~\ref{tb:ablation} shows the results. As can be observed, the significant role of the two features is highlighted by the poor performance if they are removed. 
For the ablation of the context memory, the results also demonstrate a significant degradation in performance. Without historical memory restoration, algorithms are compelled to optimize the population from scratch in each period, hindering comprehensive algorithm cooperation

\begin{table}[t]
\centering
\caption{The ablation study performance.}
\resizebox{0.65\columnwidth}{!}{%
\begin{tabular}{c|cc|cc|cc|cc}
\hline
\begin{tabular}[c]{@{}c@{}}Problem\\Class\end{tabular}
& \multicolumn{2}{c|}{RL-DAS w/o LA}
& \multicolumn{2}{c|}{RL-DAS w/o AH}
& \multicolumn{2}{c|}{RL-DAS w/o Context}
& \multicolumn{2}{c}{RL-DAS}
\\ \hline

C1
& \textbf{100.0}\% $\approx$        & 1.425e5 $-$
& \textbf{100.0}\% $\approx$        & 1.383e5 $-$
& 99.96\% $-$        & 2.000e5 $-$
& \textbf{100.0}\%        & \textbf{1.127e5}
\\
C2
& 82.29\% $-$                 & \textbf{2.000e5} $\approx$
& 82.73\% $-$                 & \textbf{2.000e5} $\approx$
& 56.34\% $-$                 & \textbf{2.000e5} $\approx$
& \textbf{85.97}\%        & \textbf{2.000e5}
\\
C3
& 95.31\% $-$                 & \textbf{2.000e5} $\approx$
& 95.36\% $-$                 & \textbf{2.000e5} $\approx$
& 90.17\% $-$                 & \textbf{2.000e5} $\approx$
& \textbf{95.63}\%        & \textbf{2.000e5}
\\
C4
& \textbf{100.0}\% $\approx$        & \textbf{2.000e5} $\approx$
& \textbf{100.0}\% $\approx$        & \textbf{2.000e5} $\approx$
& 99.96\% $-$        & \textbf{2.000e5} $\approx$
& \textbf{100.0}\%        & \textbf{2.000e5}
\\
C5
& \textbf{100.0}\% $\approx$        & 2.000e5 $-$
& \textbf{100.0}\% $\approx$        & 2.000e5 $-$
& 99.97\% $-$        & 2.000e5 $-$
& \textbf{100.0}\%        & \textbf{1.996e5}
\\
C6
& 98.96\% $-$                 & \textbf{2.000e5} $\approx$
& 98.91\% $-$                 & \textbf{2.000e5} $\approx$
& 95.08\% $-$                 & \textbf{2.000e5} $\approx$
& \textbf{99.39}\%        & \textbf{2.000e5}
\\
C7
& 99.99\% $-$                 & \textbf{2.000e5} $\approx$
& \textbf{100.0}\% $\approx$        & \textbf{2.000e5} $\approx$
& 99.21\% $-$        & \textbf{2.000e5} $\approx$
& \textbf{100.0}\%        & \textbf{2.000e5}
\\
C8
& 79.32\% $-$                 & 1.851e5 $-$
& 79.77\% $-$                 & 1.873e5 $-$
& 68.43\% $-$                 & 2.000e5 $-$
& \textbf{80.48}\%        & \textbf{1.821e5}
\\
C9
& 89.52\% $-$                 & 1.905e5 $-$
& 89.86\% $-$                 & 1.912e5 $-$
& 80.28\% $-$                 & 2.000e5 $-$
& \textbf{90.22}\%        & \textbf{1.882e5}
\\
C10
& 88.24\% $-$                 & 1.916e5 $\approx$
& 87.76\% $-$                 & 1.933e5 $-$
& 77.67\% $-$                 & 2.000e5 $-$
& \textbf{88.79}\%        & \textbf{1.914e5}
\\
C11
& 92.95\% $-$                 & 1.939e5 $-$
& 93.16\% $-$                 & 1.948e5 $-$
& 84.27\% $-$                 & 2.000e5 $-$
& \textbf{93.94}\%        & \textbf{1.903e5}
\\ \hline
& \multicolumn{2}{c|}{10 / 1 / 0}
& \multicolumn{2}{c|}{9 / 2 / 0}
& \multicolumn{2}{c|}{11 / 0 / 0}
& \multicolumn{2}{c}{- / - / -}
\\
\end{tabular}%
}
\label{tb:ablation}
\end{table}

\subsubsection{Reward Design}\label{Exp:rew}
RL-DAS adopts a reward mechanism that combines absolute cost descent and an adaptation term based on the convergence speed. 
Within this part, we thoroughly examine five different reward designs, by taking the performance of RL-DAS on 10D problems as an example.

\begin{enumerate}
    \item Absolute cost descent reward: The decline of the optimal cost of the population at time step $t$ normalized by the optimal cost of the initial population. 
    \begin{equation}
        r1 = \frac{cost^*_{t-1} - cost^*_t}{cost^*_0}
    \end{equation}
    \item Relative cost descent ratio: The optimal cost decline normalized by the optimal cost in previous generation.
    \begin{equation}
        r2 = \frac{cost^*_{t-1} - cost^*_t}{cost^*_{t-1}}
    \end{equation}
    \item Zero-one reward: The reward will be 1 if the optimal cost of the population decreased. Otherwise it is 0.
    \begin{equation}
        r3 = \begin{cases}
            1 & \text{If } cost^*_{t-1} - cost^*_t > 0 \\
            0 & \text{Otherwise}
        \end{cases}
    \end{equation}
    \item Restricted zero-one reward: A restriction is introduced to the zero-one reward that the reward will be 1 only if the difference between costs is larger than a threshold (2.5\%).
    \begin{equation}
        r4 = \begin{cases}
            1 & \text{If } \frac{cost^*_{t-1} - cost^*_t}{cost^*_{t-1}} > 0.025 \\
            0 & \text{Otherwise}
        \end{cases}
    \end{equation}
    \item Adjusted cost descent reward: Apply an external term on $r1$ to include the consumed FEs to reach the termination. It is shown in Eq.~(\ref{eq:reward}) and applied in RL-DAS.
\end{enumerate}

\begin{table}[]
\centering
\caption{The performance under different reward schemes.}
\resizebox{0.40\columnwidth}{!}{%
\begin{tabular}{c|ccccc}
\hline
 \multicolumn{1}{c|}{\multirow{2}{*}{\begin{tabular}[c]{@{}c@{}}Problem\\Class\end{tabular}}} 
 & \multicolumn{5}{c}{Reward Type}
\\ \cline{2-6}
 
& r1
& r2
& r3
& r4
& Ours
\\ \hline
\multirow{2}{*}{C1}
&\textbf{100.0}\% $\approx$     
& \textbf{100.0}\% $\approx$   
&\textbf{100.0}\% $\approx$
&\textbf{100.0}\% $\approx$
&\textbf{100.0}\%
\\
& 1.345e5 $-$
& \textbf{1.035e5} $+$
& 1.360e5 $-$
& 1.359e5 $-$
& 1.127e5
\\ \hline
\multirow{2}{*}{C2}
& 85.79\% $-$
& 85.73\% $-$
& 85.46\% $-$
& 83.96\% $-$
& \textbf{85.97}\%
\\
& \textbf{2.000e5} $\approx$
& \textbf{2.000e5} $\approx$
& \textbf{2.000e5} $\approx$
& \textbf{2.000e5} $\approx$
& \textbf{2.000e5}
\\ \hline
\multirow{2}{*}{C3}
& 95.36\% $-$
& 95.37\% $-$
& 95.33\% $-$
& 94.14\% $-$
& \textbf{95.63}\%
\\
& \textbf{2.000e5} $\approx$
& \textbf{2.000e5} $\approx$
& \textbf{2.000e5} $\approx$
& \textbf{2.000e5} $\approx$
& \textbf{2.000e5}
\\ \hline
\multirow{2}{*}{C4}
& \textbf{100.0}\% $\approx$
& \textbf{100.0}\% $\approx$
& \textbf{100.0}\% $\approx$
& \textbf{100.0}\% $\approx$
& \textbf{100.0}\%
\\
& \textbf{2.000e5} $\approx$
& \textbf{2.000e5} $\approx$
& \textbf{2.000e5} $\approx$
& \textbf{2.000e5} $\approx$
& \textbf{2.000e5}
\\ \hline
\multirow{2}{*}{C5}
& \textbf{100.0}\% $\approx$
& \textbf{100.0}\% $\approx$
& \textbf{100.0}\% $\approx$
& \textbf{100.0}\% $\approx$
& \textbf{100.0}\%
\\
& 2.000e5 $-$
& 1.999e5 $-$
& 2.000e5 $-$
& 2.000e5 $-$
& \textbf{1.996e5}
\\ \hline
\multirow{2}{*}{C6}
& 99.01\% $-$
& 98.46\% $-$
& 98.68\% $-$
& 97.91\% $-$
& \textbf{99.39}\%
\\
& \textbf{2.000e5} $\approx$
& \textbf{2.000e5} $\approx$
& \textbf{2.000e5} $\approx$
& \textbf{2.000e5} $\approx$
& \textbf{2.000e5}
\\ \hline
\multirow{2}{*}{C7}
& \textbf{100.0}\% $\approx$
& \textbf{100.0}\% $\approx$
& \textbf{100.0}\% $\approx$
& \textbf{100.0}\% $\approx$
& \textbf{100.0}\%
\\
& \textbf{2.000e5} $\approx$
& \textbf{2.000e5} $\approx$
& \textbf{2.000e5} $\approx$
& \textbf{2.000e5} $\approx$
& \textbf{2.000e5}
\\ \hline
\multirow{2}{*}{C8}
& 80.36\% $\approx$
& 80.01\% $-$
& 79.77\% $-$
& 80.43\% $\approx$
& \textbf{80.48}\%
\\
& 1.923e5 $-$
& \textbf{1.798e5} $+$
& 1.891e5 $-$
& 1.957e5 $-$
& 1.821e5
\\ \hline
\multirow{2}{*}{C9}
& \textbf{90.20}\% $\approx$
& 89.86\% $-$
& 89.73\% $-$
& 89.17\% $-$
& \textbf{90.22}\%
\\
& 1.924e5 $-$
& \textbf{1.880e5} $\approx$
& 1.916e5 $-$
& 1.943e5 $-$
& 1.882e5
\\ \hline
\multirow{2}{*}{C10}
& 88.61\% $-$
& 88.55\% $-$
& 88.16\% $-$
& 87.63\% $-$
& \textbf{88.79}\%
\\
& 1.951e5 $-$
& 1.916e5 $\approx$
& 1.943e5 $-$
& 1.949e5 $-$
& \textbf{1.914e5}
\\ \hline
\multirow{2}{*}{C11}
& 93.65\% $-$
& 93.56\% $-$
& 93.52\% $-$
& 93.07\% $-$
& \textbf{93.94}\%
\\
& 1.924e5 $-$
& \textbf{1.886e5} $+$
& 1.921e5 $-$
& 1.941e5 $-$
& 1.903e5
\\ \hline
& 9 / 2 / 0
& 8 / 2 / 1
& 9 / 2 / 0
& 9 / 2 / 0
& - / - / -
\\

\end{tabular}%
}
\label{tb:reward}
\end{table}
The comparison results for 10D problems are presented in Table~\ref{tb:reward}. Our reward explicitly considers both cost and speed, offering valuable guidance to the agent. It is evident from the table that our reward outperforms others on the majority of problems, particularly in terms of optimization speed. In contrast, alternative reward functions such as $r1$ offer no differentiation on FE consumption, resulting in identical rewards for fast or slow convergence. Consequently, the agent lacks motivation to expedite optimization, leading to relatively poor performance.

\subsubsection{Schedule Interval}

\begin{figure}[!t]
\centering
\hspace{-5mm}
\includegraphics[width=0.45\columnwidth]{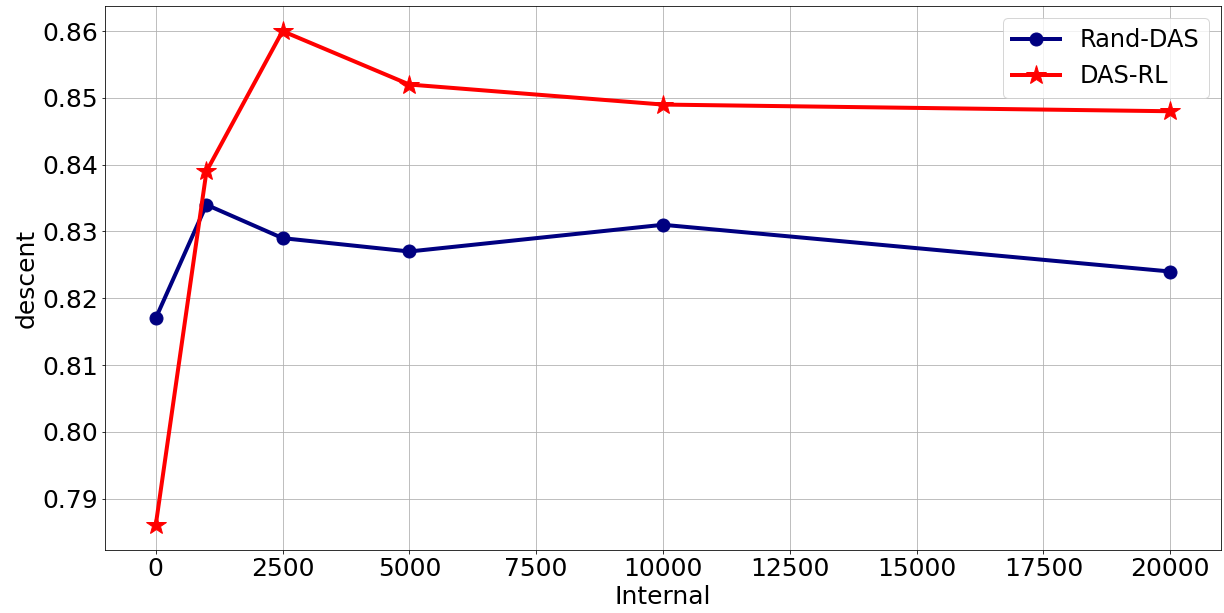}
\caption{The performance change along schedule intervals.}
\label{interval}
\end{figure}

In the previous experiment, we fixed the interval on 10D problems to 2500 FEs. However, different intervals can have an impact on performance. 
In this section, we present the performance of RL-DAS and the Rand-DAS for intervals with values of 0, 1000, 2500, 5000, 10000, and 20000 on 10D Schwefel. The ``0'' interval indicates switching at every EC generation. The performance is shown in Fig.~\ref{interval}.

The performance of RL-DAS and Rand-DAS show similar tendency.
Among these intervals, RL-DAS with 2500 interval achieves the best performance. The smallest interval, 0, presents a significantly poor performance due to the algorithm's integrity being compromised by too frequent switching and a large number of feature samplings (which consume a lot of additional evaluations). These factors make it significantly worse than Rand-DAS which has no feature samplings. The performance of intervals larger than 2500 shows a gentle decrease as they lose flexibility but retain algorithm integrity and use less feature sampling. Therefore, 
an interval of 2500 is suggested in this paper for RL-DAS.

\section{Conclusion}\label{Sec:conclude}

This paper proposed the RL-DAS framework to select the optimal algorithm at different optimization stages for given problems. We characterized the optimization state space through analyzing the fitness landscape and algorithm history. We then designed a deep neural network to infer the best action (namely, algorithm selection) for each state, which was trained in the actor-critic-based PPO manner. Meanwhile, the algorithm context memory mechanism was deployed to support the warm start of different algorithms. RL-DAS allows for seamless switching among candidate algorithms to optimize the problem in a dynamic-online fashion, while it is simple and generic and has the potential to boost many EC algorithms.

We applied the framework to a group of DE algorithms for investigation. Experimental results showed that RL-DAS not only improved the overall optimization performance but also exhibited desirable generalization ability across different problem classes. The detailed analysis also verified the importance of DAS, which provided new opportunities to boost the traditional AS method. There are still some limitations such as the manual state representation and challenging integration of complex algorithms. Therefore, considerable future efforts are warranted  to explore this area, such as automatic state representation and the inclusion of more candidate algorithms.

\newpage
\begin{appendices}
\centering\section*{APPENDIX}
\end{appendices}
\appendix
\section{Detailed Definition of LA Features}

In the main body of our paper, we briefly introduce the main idea of the 9 LA features. In this section, we present their detailed formulation:

\begin{enumerate}[label = \alph*)]
    \item $feature_1$: For the first feature measuring the value of cost, we adopt the current best cost $cost_t^*$ normalized by the best cost $cost^*_{0}$ in the initial population as an optimization progress feature:
    \begin{equation}
         feature_1 = \frac{cost_t^*}{cost^*_{0}}
    \end{equation}
   
    \item $feature_2$: The fitness distance correlation (FDC)~\cite{tomassini2005study} is a feature that describes the complexity of the problem by evaluating the relationship between fitness value and the distance of the solution from the optimum. Given a set of individuals' distance to the best individual $Dist = \{d_1^*, \cdots, d_N^*\}$ and their costs $C = \{c_1,\cdots, c_N\}$, the feature is calculated as:
    \begin{equation}
        feature_2 = \frac{\frac{1}{N}\sum_{i=1}^{N}{(c_i - \Bar{c})(d_i^* - \Bar{d}^*)}}{\delta_{Dist}\delta_C}
    \end{equation}
    where $\delta_{Dist}$ and $\delta_C$ are the variances of $Dist$ and $C$.
    \item $feature_3$: The dispersion difference~\cite{lunacek2006dispersion} measures the distribution difference between the top 10\% and the whole population: 
    \begin{equation}
        feature_3 = \Bar{d}_\text{top} - \Bar{d} 
    \end{equation}
    where $\Bar{d}_\text{top}$ is the average pairwise distance of top 10\% individuals. It is applied to analyze the funnelity of the problem landscape: a single funnel problem has a smaller dispersion difference value as the top 10\% individuals gather around the optimal point and have smaller inner distance, otherwise for the multi-funnel landscape, this value would be much larger.
    \item $feature_4$: The maximum distance among population 
    describes the convergence state of population. It is calculated by the maximum distance in the population and normalized by the diameter of the search space: 
    \begin{equation}
        feature_4=\frac{d_{\max}}{diameter}
    \end{equation}
    \item $feature_5$: The next four features are adopted to measure the evolvability of the population on the given problem. First, the negative slope coefficient (NSC)~\cite{vanneschi2004fitness} measures the difference between the costs change incurred by the population evolution and by making small tentative random walk steps. 
    In this paper, these tentative steps are made by two randomly selected algorithms from $\Lambda$.
    Given the ascending sorted evolved population cost set ($C$) and the randomly sampled population cost set ($C'$) that are both uniformly segmented into $m$ parts as $\{C_1, \cdots, C_m\}$ and $\{C'_1, \cdots, C'_m\}$, NSC is computed by:
    \begin{equation}
        feature_5 = \min\left(\sum_{i=1}^{m}{\frac{\Bar{C}_{i+1} - \Bar{C}_{i}}{\Bar{C'} _{i+1}- \Bar{C}'_{i}}}, 0\right)
    \end{equation}
    which reflects the hardness in solving the problem: 
    a high value such as 0 indicates an easy problem, whereas a negative NSC indicates a more complex problem landscape. 
    \item $feature_6$: The average neutral ratio (ANR)~\cite{vanneschi2007comprehensive} has a similar consideration of NSC, but it measures the performance changes between the current population and $S$ different populations sampled through candidate algorithms. 
    Given the cost set of sampled populations $\{C'^1, \cdots, C'^S\}$ where $C'^j = \{c^j_1, \cdots, c^j_N\}$, this feature is calculated as
    \begin{equation}
        feature_6 = \frac{\sum_i^N{\sum_j^S{|c_i - c_i^j| < eps}}}{N\cdot S}
    \end{equation}
    where $eps$ is a precision value. The higher the ANR is, the problem landscape is more rugged.
    
    \item $feature_7$: Inspired by Wang et al.~\cite{wang2017population}, we also propose the Best-Worst Improvement to measure the evolvability. The Best Improvement ($feature_7$) measures the optimization difficulty of the current population, which is calculated by 
    the ratio of sampled individuals that make no improvement 
    over the current population:
    \begin{equation}
        feature_7 = \frac{\sum_i^N{\alpha_i}}{N}
    \end{equation}
    where 
    \begin{equation}
        \alpha_i = \begin{cases}
            0&\text{if}~\left(\sum_j^S{\mathbb{I}(c_i^j < c_i)}\right) > 0\\
            1&\text{otherwise}
        \end{cases}
    \end{equation}
    Here, $\alpha_i$ is the contribution of the $i$-th individual in the current population ($c_i$ is its cost), and $c_i^j$ is the cost of the $i$-th individual in the $j$-th sampled population. The $\alpha_i$ will be 1 if the current individual $i$ is the best over the sampled individuals with small potential to be improved. 
    \item $feature_8$: Contrary to the Best Improvement, the Worst Improvement ($feature_8$) describes the optimization potential and uses the ratio of individuals that do not get worse:
    \begin{equation}
        feature_8 = \frac{\sum_i^N{\beta_i}}{N}
    \end{equation}
    where
    \begin{equation}
        \beta_i = \begin{cases}
            0&\text{if}~\left(\sum_j^S{\mathbb{I}(c_i^j > c_i})\right) < S\\
            1&\text{otherwise}
        \end{cases}
    \end{equation}
    Here, $\beta_i$ will be 1 if the individual is the worst among the sampled individuals with high potential.
    \item $feature_9$: It is beneficial to let the agent be aware of the consumption of the computational budget, so we introduce the ratio of consumed $FEs$ over the maximum as a feature:
    \begin{equation}
        feature_9 = \frac{FEs}{MaxFEs}
    \end{equation}
\end{enumerate}

\section{Experiment Settings}

The parameters of JDE21, MadDE, and NL-SHADE-RSP integrated in RL-DAS are kept the same as in their original papers. The exceptions are the parameters shared across the candidates and have different settings in different algorithms, such as the size of elite archive, parameter memories and population. To fix these gaps, we choose the maximal settings $2.3\times N$ for the archive size (following MadDE) and $20\times D$ for the memory sizes (following NL-SHADE-RSP). The population size range follows the setting of JDE21 to ensure its multi-population division, which is [30, 170]. The parameters for comparisons are kept the same as in their original papers.

The $MaxFEs$ is set to 200,000 for 10D problems and 1,000,000 for 20D problems following the conventions of CEC2021 competition (note that the FEs costed in the fitness landscape analysis for state representation of RL-DAS are considered for a fair comparison). The schedule interval $\varDelta$ is set to 2,500 for 10D and 8,000 for 20D problems. During training, the instance sets are batched in the size of 16. The agent learns for 200 epochs with the learning rate $\lambda\!=\!10^{-5}$ for both of the actor and the critic. For the testing, each algorithm is executed 30 runs for each test instance. The algorithm is terminated when either the $MaxFEs$ is exhausted or the best cost is lower than the termination error $10^{-8}$.

\begin{table*}[t]
\centering
\caption{
Comparing RL-DAS with comparison algorithms on Shifted \& Rotated 10D problems. }
\resizebox{0.95\textwidth}{!}{%
\begin{tabular}{cc|cc|ccccccccccc}
\hline
  \multicolumn{2}{c|}{\multirow{3}{*}{Taxonomy}} 
& \multicolumn{2}{c|}{\multirow{3}{*}{Algorithm}} 
& \multicolumn{11}{c}{\multirow{1}{*}{Problem Class}}
\\ \cline{5-15}
 & & &
 & \multicolumn{1}{l}{\multirow{2}{*}{\quad C1}}
 & \multicolumn{1}{l}{\multirow{2}{*}{\quad C2}}
 & \multicolumn{1}{l}{\multirow{2}{*}{\quad C3}}
 & \multicolumn{1}{l}{\multirow{2}{*}{\quad C4}}
 & \multicolumn{1}{l}{\multirow{2}{*}{\quad C5}}
 & \multicolumn{1}{l}{\multirow{2}{*}{\quad C6}}
 & \multicolumn{1}{l}{\multirow{2}{*}{\quad C7}}
 & \multicolumn{1}{l}{\multirow{2}{*}{\quad C8}}
 & \multicolumn{1}{l}{\multirow{2}{*}{\quad C9}}
 & \multicolumn{1}{l}{\multirow{2}{*}{\quad C10}}
 & \multicolumn{1}{l}{\multirow{2}{*}{\quad C11}}
 \\
 & & & & & & & & & & & & & &
 \\ \hline
\multicolumn{2}{c|}{\multirow{3}{*}{DAS}}
& \textbf{RL-DAS}
& \multicolumn{1}{|c|}{\begin{tabular}[c]{@{}c@{}}Mean\\Std\end{tabular}}
& \begin{tabular}[c]{@{}c@{}}\textbf{2.974e-8}\\\textbf{2.156e-8}\end{tabular}
& \begin{tabular}[c]{@{}c@{}}\textbf{3.334e2}\\\textbf{1.108e2}\end{tabular}
& \begin{tabular}[c]{@{}c@{}}\textbf{1.738e1}\\\textbf{1.666e0}\end{tabular}
& \begin{tabular}[c]{@{}c@{}}9.982e-1\\3.487e-1\end{tabular}
& \begin{tabular}[c]{@{}c@{}}\textbf{1.587e1}\\\textbf{2.314e1}\end{tabular}
& \begin{tabular}[c]{@{}c@{}}\textbf{5.659e0}\\\textbf{3.211e0}\end{tabular}
& \begin{tabular}[c]{@{}c@{}}\textbf{9.783e0}\\\textbf{1.058e1}\end{tabular}
& \begin{tabular}[c]{@{}c@{}}\textbf{5.192e1}\\\textbf{1.259e1}\end{tabular}
& \begin{tabular}[c]{@{}c@{}}\textbf{7.988e1}\\\textbf{2.468e1}\end{tabular}
& \begin{tabular}[c]{@{}c@{}}1.715e2\\1.501e1\end{tabular}
& \begin{tabular}[c]{@{}c@{}}\textbf{7.386e1}\\\textbf{2.364e1}\end{tabular}
\\
&
& \textbf{Rand-DAS}
& \multicolumn{1}{|c|}{\begin{tabular}[c]{@{}c@{}}Mean\\Std\end{tabular}}
& \begin{tabular}[c]{@{}c@{}}6.473e-8           \\3.468e-7\end{tabular}
& \begin{tabular}[c]{@{}c@{}}4.071e2            \\1.546e2\end{tabular}
& \begin{tabular}[c]{@{}c@{}}1.828e1            \\2.342e0\end{tabular}
& \begin{tabular}[c]{@{}c@{}}1.113e0   \\4.679e-1\end{tabular}
& \begin{tabular}[c]{@{}c@{}}1.754e1   \\3.178e1\end{tabular}
& \begin{tabular}[c]{@{}c@{}}6.216e0            \\4.156e0\end{tabular}
& \begin{tabular}[c]{@{}c@{}}9.997e0   \\1.189e1\end{tabular}
& \begin{tabular}[c]{@{}c@{}}5.355e1            \\1.673e1\end{tabular}
& \begin{tabular}[c]{@{}c@{}}8.242e1            \\3.479e1\end{tabular}
& \begin{tabular}[c]{@{}c@{}}1.757e2            \\1.712e1\end{tabular}
& \begin{tabular}[c]{@{}c@{}}7.691e1            \\2.856e1\end{tabular}
\\ \hline
\multicolumn{2}{c|}{\multirow{1}{*}{AS}}
& \textbf{AS$^*$}
& \multicolumn{1}{|c|}{\begin{tabular}[c]{@{}c@{}}Mean\\Std\end{tabular}}
& \begin{tabular}[c]{@{}c@{}}5.424e-8           \\1.867e-9\end{tabular}
& \begin{tabular}[c]{@{}c@{}}3.415e2            \\1.068e2\end{tabular}
& \begin{tabular}[c]{@{}c@{}}1.833e1            \\2.153e0\end{tabular}
& \begin{tabular}[c]{@{}c@{}}\textbf{9.546e-1}  \\\textbf{3.681e-1}\end{tabular}
& \begin{tabular}[c]{@{}c@{}}1.628e1  \\2.610e1\end{tabular}
& \begin{tabular}[c]{@{}c@{}}6.123e0            \\3.798e0\end{tabular}
& \begin{tabular}[c]{@{}c@{}}9.817e0   \\9.998e0\end{tabular}
& \begin{tabular}[c]{@{}c@{}}5.357e1            \\1.351e1\end{tabular}
& \begin{tabular}[c]{@{}c@{}}8.217e1            \\2.754e1\end{tabular}
& \begin{tabular}[c]{@{}c@{}}1.752e2            \\1.216e1\end{tabular}
& \begin{tabular}[c]{@{}c@{}}7.668e1            \\2.234e1\end{tabular}
\\ \hline
\multicolumn{2}{c|}{\multirow{5}{*}{Backbones}}
& \textbf{JDE21}
& \multicolumn{1}{|c|}{\begin{tabular}[c]{@{}c@{}}Mean\\Std\end{tabular}}
& \begin{tabular}[c]{@{}c@{}}8.675e-7           \\1.122e-6\end{tabular}
& \begin{tabular}[c]{@{}c@{}}6.292e2            \\1.923e5\end{tabular}
& \begin{tabular}[c]{@{}c@{}}2.005e1            \\2.531e0\end{tabular}
& \begin{tabular}[c]{@{}c@{}}1.102e0   \\3.648e-1\end{tabular}
& \begin{tabular}[c]{@{}c@{}}1.273e2            \\2.354e2\end{tabular}
& \begin{tabular}[c]{@{}c@{}}9.185e0            \\5.254e0\end{tabular}
& \begin{tabular}[c]{@{}c@{}}2.918e1            \\3.047e1\end{tabular}
& \begin{tabular}[c]{@{}c@{}}5.650e1            \\1.276e1\end{tabular}
& \begin{tabular}[c]{@{}c@{}}9.418e1            \\2.317e1\end{tabular}
& \begin{tabular}[c]{@{}c@{}}1.771e2            \\1.056e1\end{tabular}
& \begin{tabular}[c]{@{}c@{}}9.727e1            \\3.179e1\end{tabular}
\\
&
& \textbf{MadDE}
& \multicolumn{1}{|c|}{\begin{tabular}[c]{@{}c@{}}Mean\\Std\end{tabular}}
& \begin{tabular}[c]{@{}c@{}}4.162e-8  \\4.311e-8\end{tabular}
& \begin{tabular}[c]{@{}c@{}}3.888e2            \\1.135e2\end{tabular}
& \begin{tabular}[c]{@{}c@{}}1.834e1            \\2.649e0\end{tabular}
& \begin{tabular}[c]{@{}c@{}}9.757e-1  \\4.17e-1\end{tabular}
& \begin{tabular}[c]{@{}c@{}}1.834e1   \\2.228e1\end{tabular}
& \begin{tabular}[c]{@{}c@{}}6.123e0            \\3.105e0\end{tabular}
& \begin{tabular}[c]{@{}c@{}}1.018e1   \\1.086e1\end{tabular}
& \begin{tabular}[c]{@{}c@{}}5.578e1            \\1.233e1\end{tabular}
& \begin{tabular}[c]{@{}c@{}}8.356e1            \\3.670e1\end{tabular}
& \begin{tabular}[c]{@{}c@{}}1.847e2            \\1.168e1\end{tabular}
& \begin{tabular}[c]{@{}c@{}}7.667e1            \\2.305e1\end{tabular}
\\
&
& \textbf{NL-SHADE-RSP}
& \multicolumn{1}{|c|}{\begin{tabular}[c]{@{}c@{}}Mean\\Std\end{tabular}}
& \begin{tabular}[c]{@{}c@{}}7.892e-4           \\8.697e-3\end{tabular}
& \begin{tabular}[c]{@{}c@{}}4.577e2            \\1.356e2\end{tabular}
& \begin{tabular}[c]{@{}c@{}}1.901e1            \\2.462e0\end{tabular}
& \begin{tabular}[c]{@{}c@{}}9.810e-1  \\3.471e-1\end{tabular}
& \begin{tabular}[c]{@{}c@{}}1.781e1   \\2.657e1\end{tabular}
& \begin{tabular}[c]{@{}c@{}}2.097e1            \\1.348e1\end{tabular}
& \begin{tabular}[c]{@{}c@{}}1.135e1   \\1.218e1\end{tabular}
& \begin{tabular}[c]{@{}c@{}}5.477e1            \\1.576e1\end{tabular}
& \begin{tabular}[c]{@{}c@{}}8.258e1            \\3.971e1\end{tabular}
& \begin{tabular}[c]{@{}c@{}}1.766e2            \\1.210e1\end{tabular}
& \begin{tabular}[c]{@{}c@{}}8.703e1            \\2.897e1\end{tabular}
\\ \hline
\multicolumn{1}{c}{\multirow{9}{*}{\begin{tabular}[c]{@{}c@{}}Other\\Advanced\\DE\\Variants\end{tabular}}}
& \multicolumn{1}{|c|}{\multirow{1}{*}{Adaptive}}
& \textbf{EPSDE}
& \multicolumn{1}{|c|}{\begin{tabular}[c]{@{}c@{}}Mean\\Std\end{tabular}}
& \begin{tabular}[c]{@{}c@{}}1.789e-7           \\2.456e-7\end{tabular}
& \begin{tabular}[c]{@{}c@{}}5.767e2             \\2.015e2\end{tabular}
& \begin{tabular}[c]{@{}c@{}}1.780e1             \\2.049e0\end{tabular}
& \begin{tabular}[c]{@{}c@{}}1.072e0    \\4.077e-1\end{tabular}
& \begin{tabular}[c]{@{}c@{}}2.047e1    \\3.542e1\end{tabular}
& \begin{tabular}[c]{@{}c@{}}7.144e0             \\4.351e0\end{tabular}
& \begin{tabular}[c]{@{}c@{}}1.105e1    \\1.089e0\end{tabular}
& \begin{tabular}[c]{@{}c@{}}6.267e1             \\1.437e1\end{tabular}
& \begin{tabular}[c]{@{}c@{}}1.045e2             \\3.688e1\end{tabular}
& \begin{tabular}[c]{@{}c@{}}1.502e2             \\9.547e0\end{tabular}
& \begin{tabular}[c]{@{}c@{}}9.227e1             \\3.151e1\end{tabular}
\\ \cline{2-15}
& \multicolumn{1}{|c|}{\multirow{1}{*}{Ensemble}}
& \textbf{EDEV}
& \multicolumn{1}{|c|}{\begin{tabular}[c]{@{}c@{}}Mean\\Std\end{tabular}}
& \begin{tabular}[c]{@{}c@{}}2.015e-7           \\2.346e-7\end{tabular}
& \begin{tabular}[c]{@{}c@{}}5.358e2             \\1.713e2\end{tabular}
& \begin{tabular}[c]{@{}c@{}}1.778e1             \\1.941e0\end{tabular}
& \begin{tabular}[c]{@{}c@{}}9.954e-1   \\3.178e-1\end{tabular}
& \begin{tabular}[c]{@{}c@{}}2.258e1    \\2.985e1\end{tabular}
& \begin{tabular}[c]{@{}c@{}}6.772e0             \\3.789e0\end{tabular}
& \begin{tabular}[c]{@{}c@{}}1.096e1    \\1.131e1\end{tabular}
& \begin{tabular}[c]{@{}c@{}}6.020e1             \\1.476e1\end{tabular}
& \begin{tabular}[c]{@{}c@{}}1.009e2             \\2.977e1\end{tabular}
& \begin{tabular}[c]{@{}c@{}}\textbf{1.368e2}    \\\textbf{9.471e0}\end{tabular}
& \begin{tabular}[c]{@{}c@{}}8.508e1             \\2.777e1\end{tabular}
\\ \cline{2-15} 
& \multicolumn{1}{|c|}{\multirow{5}{*}{\begin{tabular}[c]{@{}c@{}}RL\\Assisted\end{tabular}}}
& \textbf{DEDQN}
& \multicolumn{1}{|c|}{\begin{tabular}[c]{@{}c@{}}Mean\\Std\end{tabular}}
& \begin{tabular}[c]{@{}c@{}}6.492e3             \\7.279e3\end{tabular}
& \begin{tabular}[c]{@{}c@{}}6.737e2             \\2.346e2\end{tabular}
& \begin{tabular}[c]{@{}c@{}}3.011e1             \\3.468e0\end{tabular}
& \begin{tabular}[c]{@{}c@{}}1.036e0    \\3.017e-1\end{tabular}
& \begin{tabular}[c]{@{}c@{}}1.366e2    \\1.791e2\end{tabular}
& \begin{tabular}[c]{@{}c@{}}1.203e2             \\5.671e1\end{tabular}
& \begin{tabular}[c]{@{}c@{}}2.987e1    \\3.154e1\end{tabular}
& \begin{tabular}[c]{@{}c@{}}6.360e1             \\1.279e1\end{tabular}
& \begin{tabular}[c]{@{}c@{}}1.354e2             \\2.057e1\end{tabular}
& \begin{tabular}[c]{@{}c@{}}1.859e2             \\1.636e1\end{tabular}
& \begin{tabular}[c]{@{}c@{}}1.053e2             \\3.781e1\end{tabular}
\\
& \multicolumn{1}{|c|}{}
& \textbf{DEDDQN}
& \multicolumn{1}{|c|}{\begin{tabular}[c]{@{}c@{}}Mean\\Std\end{tabular}}
& \begin{tabular}[c]{@{}c@{}}5.271e3             \\6.342e3\end{tabular}
& \begin{tabular}[c]{@{}c@{}}6.658e2             \\1.946e2\end{tabular}
& \begin{tabular}[c]{@{}c@{}}2.701e0             \\2.978e0\end{tabular}
& \begin{tabular}[c]{@{}c@{}}1.076e0    \\2.814e-1\end{tabular}
& \begin{tabular}[c]{@{}c@{}}1.321e2    \\2.457e2\end{tabular}
& \begin{tabular}[c]{@{}c@{}}1.041e2             \\5.547e1\end{tabular}
& \begin{tabular}[c]{@{}c@{}}2.541e1    \\2.454e1\end{tabular}
& \begin{tabular}[c]{@{}c@{}}6.158e1             \\1.219e1\end{tabular}
& \begin{tabular}[c]{@{}c@{}}1.264e2             \\2.153e1\end{tabular}
& \begin{tabular}[c]{@{}c@{}}1.826e2             \\1.713e1\end{tabular}
& \begin{tabular}[c]{@{}c@{}}1.065e2             \\3.687e1\end{tabular}
\\
& \multicolumn{1}{|c|}{}
& \textbf{LDE}
& \multicolumn{1}{|c|}{\begin{tabular}[c]{@{}c@{}}Mean\\Std\end{tabular}}
& \begin{tabular}[c]{@{}c@{}}2.152e1             \\6.153e1\end{tabular}
& \begin{tabular}[c]{@{}c@{}}5.598e2             \\1.331e2\end{tabular}
& \begin{tabular}[c]{@{}c@{}}1.833e1             \\1.798e0\end{tabular}
& \begin{tabular}[c]{@{}c@{}}9.897e-1           \\2.823e-1\end{tabular}
& \begin{tabular}[c]{@{}c@{}}1.835e1            \\2.847e1\end{tabular}
& \begin{tabular}[c]{@{}c@{}}1.011e1             \\5.133e0\end{tabular}
& \begin{tabular}[c]{@{}c@{}}1.215e1            \\1.200e1\end{tabular}
& \begin{tabular}[c]{@{}c@{}}6.057e1             \\1.768e1\end{tabular}
& \begin{tabular}[c]{@{}c@{}}1.328e2             \\3.336e1\end{tabular}
& \begin{tabular}[c]{@{}c@{}}1.793e2             \\1.193e1\end{tabular}
& \begin{tabular}[c]{@{}c@{}}9.507e1             \\3.044e1\end{tabular}
\\ 
\hline
\end{tabular}
}
\label{appx:r+s}
\end{table*}
\begin{table*}[t]
\centering
\caption{
Comparing RL-DAS with comparison algorithms on Shifted \& Rotated 20D problems.}
\resizebox{0.95\textwidth}{!}{%
\begin{tabular}{cc|cc|ccccccccccc}
\hline
  \multicolumn{2}{c|}{\multirow{3}{*}{Taxonomy}} 
& \multicolumn{2}{c|}{\multirow{3}{*}{Algorithm}} 
& \multicolumn{11}{c}{\multirow{1}{*}{Problem Class}}
\\ \cline{5-15}
 & & &
 & \multicolumn{1}{l}{\multirow{2}{*}{\quad C1}}
 & \multicolumn{1}{l}{\multirow{2}{*}{\quad C2}}
 & \multicolumn{1}{l}{\multirow{2}{*}{\quad C3}}
 & \multicolumn{1}{l}{\multirow{2}{*}{\quad C4}}
 & \multicolumn{1}{l}{\multirow{2}{*}{\quad C5}}
 & \multicolumn{1}{l}{\multirow{2}{*}{\quad C6}}
 & \multicolumn{1}{l}{\multirow{2}{*}{\quad C7}}
 & \multicolumn{1}{l}{\multirow{2}{*}{\quad C8}}
 & \multicolumn{1}{l}{\multirow{2}{*}{\quad C9}}
 & \multicolumn{1}{l}{\multirow{2}{*}{\quad C10}}
 & \multicolumn{1}{l}{\multirow{2}{*}{\quad C11}}
 \\
 & & & & & & & & & & & & & &
 \\ \hline
\multicolumn{2}{c|}{\multirow{3}{*}{DAS}}
& \textbf{RL-DAS}
& \multicolumn{1}{|c|}{\begin{tabular}[c]{@{}c@{}}Mean\\Std\end{tabular}}
& \begin{tabular}[c]{@{}c@{}}\textbf{2.289e-8}\\\textbf{6.357e-8}\end{tabular}
& \begin{tabular}[c]{@{}c@{}}\textbf{1.385e3}\\\textbf{3.482e2}\end{tabular}
& \begin{tabular}[c]{@{}c@{}}\textbf{4.392e1}\\\textbf{5.346e0}\end{tabular}
& \begin{tabular}[c]{@{}c@{}}\textbf{2.034e0}\\\textbf{8.548e-1}\end{tabular}
& \begin{tabular}[c]{@{}c@{}}\textbf{2.012e2}\\\textbf{5.841e1}\end{tabular}
& \begin{tabular}[c]{@{}c@{}}\textbf{5.342e1}\\\textbf{2.104e1}\end{tabular}
& \begin{tabular}[c]{@{}c@{}}\textbf{6.079e1}\\\textbf{3.159e1}\end{tabular}
& \begin{tabular}[c]{@{}c@{}}\textbf{5.631e1}\\\textbf{3.111e1}\end{tabular}
& \begin{tabular}[c]{@{}c@{}}\textbf{1.333e2}\\\textbf{4.512e1}\end{tabular}
& \begin{tabular}[c]{@{}c@{}}3.102e2\\4.898e1\end{tabular}
& \begin{tabular}[c]{@{}c@{}}\textbf{2.302e2}\\\textbf{6.101e1}\end{tabular}
\\
&
& \textbf{Rand-DAS}
& \multicolumn{1}{|c|}{\begin{tabular}[c]{@{}c@{}}Mean\\Std\end{tabular}}
& \begin{tabular}[c]{@{}c@{}}1.087e-3           \\1.247e-3\end{tabular}
& \begin{tabular}[c]{@{}c@{}}1.553e3            \\3.678e2\end{tabular}
& \begin{tabular}[c]{@{}c@{}}4.477e1            \\5.752e0\end{tabular}
& \begin{tabular}[c]{@{}c@{}}2.623e0            \\1.504e0\end{tabular}
& \begin{tabular}[c]{@{}c@{}}2.473e2            \\9.493e1\end{tabular}
& \begin{tabular}[c]{@{}c@{}}8.921e1            \\5.189e1\end{tabular}
& \begin{tabular}[c]{@{}c@{}}9.983e1            \\5.498e1\end{tabular}
& \begin{tabular}[c]{@{}c@{}}5.756e1            \\4.199e1\end{tabular}
& \begin{tabular}[c]{@{}c@{}}1.388e2            \\5.647e1\end{tabular}
& \begin{tabular}[c]{@{}c@{}}3.180e2            \\6.677e1\end{tabular}
& \begin{tabular}[c]{@{}c@{}}2.470e2            \\7.122e1\end{tabular}
\\ \hline
\multicolumn{2}{c|}{\multirow{1}{*}{AS}}
& \textbf{AS$^*$}
& \multicolumn{1}{|c|}{\begin{tabular}[c]{@{}c@{}}Mean\\Std\end{tabular}}
& \begin{tabular}[c]{@{}c@{}}7.786e-8           \\6.871e-8\end{tabular}
& \begin{tabular}[c]{@{}c@{}}1.425e3            \\2.955e2\end{tabular}
& \begin{tabular}[c]{@{}c@{}}4.957e1            \\5.014e0\end{tabular}
& \begin{tabular}[c]{@{}c@{}}2.298e0            \\1.114e0\end{tabular}
& \begin{tabular}[c]{@{}c@{}}2.166e2            \\7.112e1\end{tabular}
& \begin{tabular}[c]{@{}c@{}}6.083e1            \\2.548e1\end{tabular}
& \begin{tabular}[c]{@{}c@{}}6.584e1            \\3.088e1\end{tabular}
& \begin{tabular}[c]{@{}c@{}}5.833e1            \\3.657e1\end{tabular}
& \begin{tabular}[c]{@{}c@{}}1.378e2            \\3.044e1\end{tabular}
& \begin{tabular}[c]{@{}c@{}}3.207e2            \\3.664e1\end{tabular}
& \begin{tabular}[c]{@{}c@{}}2.399e2            \\6.155e1\end{tabular}
\\ \hline
\multicolumn{2}{c|}{\multirow{5}{*}{Backbones}}
& \textbf{JDE21}
& \multicolumn{1}{|c|}{\begin{tabular}[c]{@{}c@{}}Mean\\Std\end{tabular}}
& \begin{tabular}[c]{@{}c@{}}8.903e-1           \\9.368e-1\end{tabular}
& \begin{tabular}[c]{@{}c@{}}2.201e3            \\4.158e2\end{tabular}
& \begin{tabular}[c]{@{}c@{}}5.310e1            \\4.955e0\end{tabular}
& \begin{tabular}[c]{@{}c@{}}4.627e0            \\2.015e0\end{tabular}
& \begin{tabular}[c]{@{}c@{}}3.539e2            \\1.292e2\end{tabular}
& \begin{tabular}[c]{@{}c@{}}1.265e2            \\5.448e1\end{tabular}
& \begin{tabular}[c]{@{}c@{}}1.463e2            \\6.912e1\end{tabular}
& \begin{tabular}[c]{@{}c@{}}6.147e1            \\4.694e1\end{tabular}
& \begin{tabular}[c]{@{}c@{}}1.440e2            \\2.108e1\end{tabular}
& \begin{tabular}[c]{@{}c@{}}3.229e2            \\2.514e1\end{tabular}
& \begin{tabular}[c]{@{}c@{}}2.951e2            \\7.624e1\end{tabular}
\\
&
& \textbf{MadDE}
& \multicolumn{1}{|c|}{\begin{tabular}[c]{@{}c@{}}Mean\\Std\end{tabular}}
& \begin{tabular}[c]{@{}c@{}}8.529e-8           \\7.156e-8\end{tabular}
& \begin{tabular}[c]{@{}c@{}}1.431e3            \\2.948e2\end{tabular}
& \begin{tabular}[c]{@{}c@{}}4.900e1            \\4.862e0\end{tabular}
& \begin{tabular}[c]{@{}c@{}}2.551e0            \\1.151e-1\end{tabular}
& \begin{tabular}[c]{@{}c@{}}2.333e2            \\7.745e1\end{tabular}
& \begin{tabular}[c]{@{}c@{}}6.083e1            \\3.161e1\end{tabular}
& \begin{tabular}[c]{@{}c@{}}7.094e1            \\3.318e1\end{tabular}
& \begin{tabular}[c]{@{}c@{}}6.012e1            \\3.991e1\end{tabular}
& \begin{tabular}[c]{@{}c@{}}1.428e2            \\1.983e1\end{tabular}
& \begin{tabular}[c]{@{}c@{}}3.316e2            \\2.167e1\end{tabular}
& \begin{tabular}[c]{@{}c@{}}2.376e2            \\5.595e1\end{tabular}
\\
&
& \textbf{NL-SHADE-RSP}
& \multicolumn{1}{|c|}{\begin{tabular}[c]{@{}c@{}}Mean\\Std\end{tabular}}
& \begin{tabular}[c]{@{}c@{}}5.489e0            \\6.017e0\end{tabular}
& \begin{tabular}[c]{@{}c@{}}1.738e3            \\3.481e2\end{tabular}
& \begin{tabular}[c]{@{}c@{}}6.256e1            \\5.515e0\end{tabular}
& \begin{tabular}[c]{@{}c@{}}7.921e0            \\3.100e0\end{tabular}
& \begin{tabular}[c]{@{}c@{}}3.752e2            \\1.519e2\end{tabular}
& \begin{tabular}[c]{@{}c@{}}1.805e2            \\8.575e1\end{tabular}
& \begin{tabular}[c]{@{}c@{}}1.737e2            \\8.321e1\end{tabular}
& \begin{tabular}[c]{@{}c@{}}6.310e1            \\4.561e1\end{tabular}
& \begin{tabular}[c]{@{}c@{}}1.461e2            \\1.686e1\end{tabular}
& \begin{tabular}[c]{@{}c@{}}3.821e2            \\3.818e1\end{tabular}
& \begin{tabular}[c]{@{}c@{}}2.870e2            \\7.177e1\end{tabular}
\\ \hline
\multicolumn{1}{c}{\multirow{9}{*}{\begin{tabular}[c]{@{}c@{}}Other\\Advanced\\DE\\Variants\end{tabular}}}
& \multicolumn{1}{|c|}{\multirow{1}{*}{Adaptive}}
& \textbf{EPSDE}
& \multicolumn{1}{|c|}{\begin{tabular}[c]{@{}c@{}}Mean\\Std\end{tabular}}
& \begin{tabular}[c]{@{}c@{}}4.320e-5           \\4.677e-5\end{tabular}
& \begin{tabular}[c]{@{}c@{}}2.107e3            \\3.982e2\end{tabular}
& \begin{tabular}[c]{@{}c@{}}5.112e1            \\4.844e0\end{tabular}
& \begin{tabular}[c]{@{}c@{}}3.630e0            \\2.554e0\end{tabular}
& \begin{tabular}[c]{@{}c@{}}3.647e2            \\1.223e2\end{tabular}
& \begin{tabular}[c]{@{}c@{}}7.055e1            \\3.149e1\end{tabular}
& \begin{tabular}[c]{@{}c@{}}1.147e2            \\5.516e1\end{tabular}
& \begin{tabular}[c]{@{}c@{}}5.841e1            \\3.694e1\end{tabular}
& \begin{tabular}[c]{@{}c@{}}1.450e2            \\3.419e1\end{tabular}
& \begin{tabular}[c]{@{}c@{}}3.038e2            \\3.422e1\end{tabular}
& \begin{tabular}[c]{@{}c@{}}2.738e2            \\6.984e1\end{tabular}
\\ \cline{2-15}
& \multicolumn{1}{|c|}{\multirow{1}{*}{Ensemble}}
& \textbf{EDEV}
& \multicolumn{1}{|c|}{\begin{tabular}[c]{@{}c@{}}Mean\\Std\end{tabular}}
& \begin{tabular}[c]{@{}c@{}}1.087e-5           \\1.583e-5\end{tabular}
& \begin{tabular}[c]{@{}c@{}}1.998e3            \\3.492e2\end{tabular}
& \begin{tabular}[c]{@{}c@{}}5.041e1            \\5.344e0\end{tabular}
& \begin{tabular}[c]{@{}c@{}}3.144e0            \\1.949e0\end{tabular}
& \begin{tabular}[c]{@{}c@{}}3.580e2            \\1.323e2\end{tabular}
& \begin{tabular}[c]{@{}c@{}}6.850e1            \\3.240e1\end{tabular}
& \begin{tabular}[c]{@{}c@{}}1.205e2            \\5.794e1\end{tabular}
& \begin{tabular}[c]{@{}c@{}}5.869e1            \\3.741e1\end{tabular}
& \begin{tabular}[c]{@{}c@{}}1.439e2            \\3.947e1\end{tabular}
& \begin{tabular}[c]{@{}c@{}}\textbf{3.032e5}   \\\textbf{3.632e1}\end{tabular}
& \begin{tabular}[c]{@{}c@{}}2.796e2            \\6.888e1\end{tabular}
\\ \cline{2-15}
& \multicolumn{1}{|c|}{\multirow{5}{*}{\begin{tabular}[c]{@{}c@{}}RL\\Assisted\end{tabular}}}
& \textbf{DEDQN}
& \multicolumn{1}{|c|}{\begin{tabular}[c]{@{}c@{}}Mean\\Std\end{tabular}}
& \begin{tabular}[c]{@{}c@{}}7.781e0            \\8.975e0\end{tabular}
& \begin{tabular}[c]{@{}c@{}}2.141e3            \\3.833e2\end{tabular}
& \begin{tabular}[c]{@{}c@{}}6.524e1            \\7.589e0\end{tabular}
& \begin{tabular}[c]{@{}c@{}}5.465e0            \\3.697e0\end{tabular}
& \begin{tabular}[c]{@{}c@{}}5.713e2            \\2.642e2\end{tabular}
& \begin{tabular}[c]{@{}c@{}}9.943e1            \\5.947e1\end{tabular}
& \begin{tabular}[c]{@{}c@{}}1.780e2            \\1.009e2\end{tabular}
& \begin{tabular}[c]{@{}c@{}}6.856e1            \\4.198e1\end{tabular}
& \begin{tabular}[c]{@{}c@{}}1.549e2            \\7.165e1\end{tabular}
& \begin{tabular}[c]{@{}c@{}}3.625e2            \\7.548e1\end{tabular}
& \begin{tabular}[c]{@{}c@{}}2.938e2            \\7.171e1\end{tabular}
\\
& \multicolumn{1}{|c|}{}
& \textbf{DEDDQN}
& \multicolumn{1}{|c|}{\begin{tabular}[c]{@{}c@{}}Mean\\Std\end{tabular}}
& \begin{tabular}[c]{@{}c@{}}6.513e0            \\8.147e0\end{tabular}
& \begin{tabular}[c]{@{}c@{}}2.232e3            \\4.895e2\end{tabular}
& \begin{tabular}[c]{@{}c@{}}6.849e1            \\8.687e0\end{tabular}
& \begin{tabular}[c]{@{}c@{}}4.874e0            \\3.211e0\end{tabular}
& \begin{tabular}[c]{@{}c@{}}5.551e2            \\2.156e2\end{tabular}
& \begin{tabular}[c]{@{}c@{}}8.844e1            \\4.969e1\end{tabular}
& \begin{tabular}[c]{@{}c@{}}1.686e2            \\9.699e1\end{tabular}
& \begin{tabular}[c]{@{}c@{}}6.647e1            \\3.669e1\end{tabular}
& \begin{tabular}[c]{@{}c@{}}1.495e2            \\6.558e1\end{tabular}
& \begin{tabular}[c]{@{}c@{}}3.624e2            \\7.794e1\end{tabular}
& \begin{tabular}[c]{@{}c@{}}2.889e2            \\7.089e1\end{tabular}
\\
& \multicolumn{1}{|c|}{}
& \textbf{LDE}
& \multicolumn{1}{|c|}{\begin{tabular}[c]{@{}c@{}}Mean\\Std\end{tabular}}
& \begin{tabular}[c]{@{}c@{}}6.498e-4           \\6.954e-4\end{tabular}
& \begin{tabular}[c]{@{}c@{}}2.289e3            \\3.161e2\end{tabular}
& \begin{tabular}[c]{@{}c@{}}5.705e1            \\6.939e0\end{tabular}
& \begin{tabular}[c]{@{}c@{}}3.211e0            \\1.941e0\end{tabular}
& \begin{tabular}[c]{@{}c@{}}4.301e2            \\1.982e2\end{tabular}
& \begin{tabular}[c]{@{}c@{}}6.288e1            \\4.177e1\end{tabular}
& \begin{tabular}[c]{@{}c@{}}9.654e1            \\5.900e1\end{tabular}
& \begin{tabular}[c]{@{}c@{}}6.305e1            \\3.766e1\end{tabular}
& \begin{tabular}[c]{@{}c@{}}1.466e2            \\7.556e1\end{tabular}
& \begin{tabular}[c]{@{}c@{}}3.471e2            \\6.999e1\end{tabular}
& \begin{tabular}[c]{@{}c@{}}2.799e2            \\7.184e1\end{tabular}
\\ 
\hline
\end{tabular}
}
\label{appx:20d}
\end{table*}

\section{In-Depth Analysis of DAS}
\begin{figure}[t]
\centering
        \subfloat[]{
        \label{schwefel_curve}
		\includegraphics[width=0.4\textwidth]{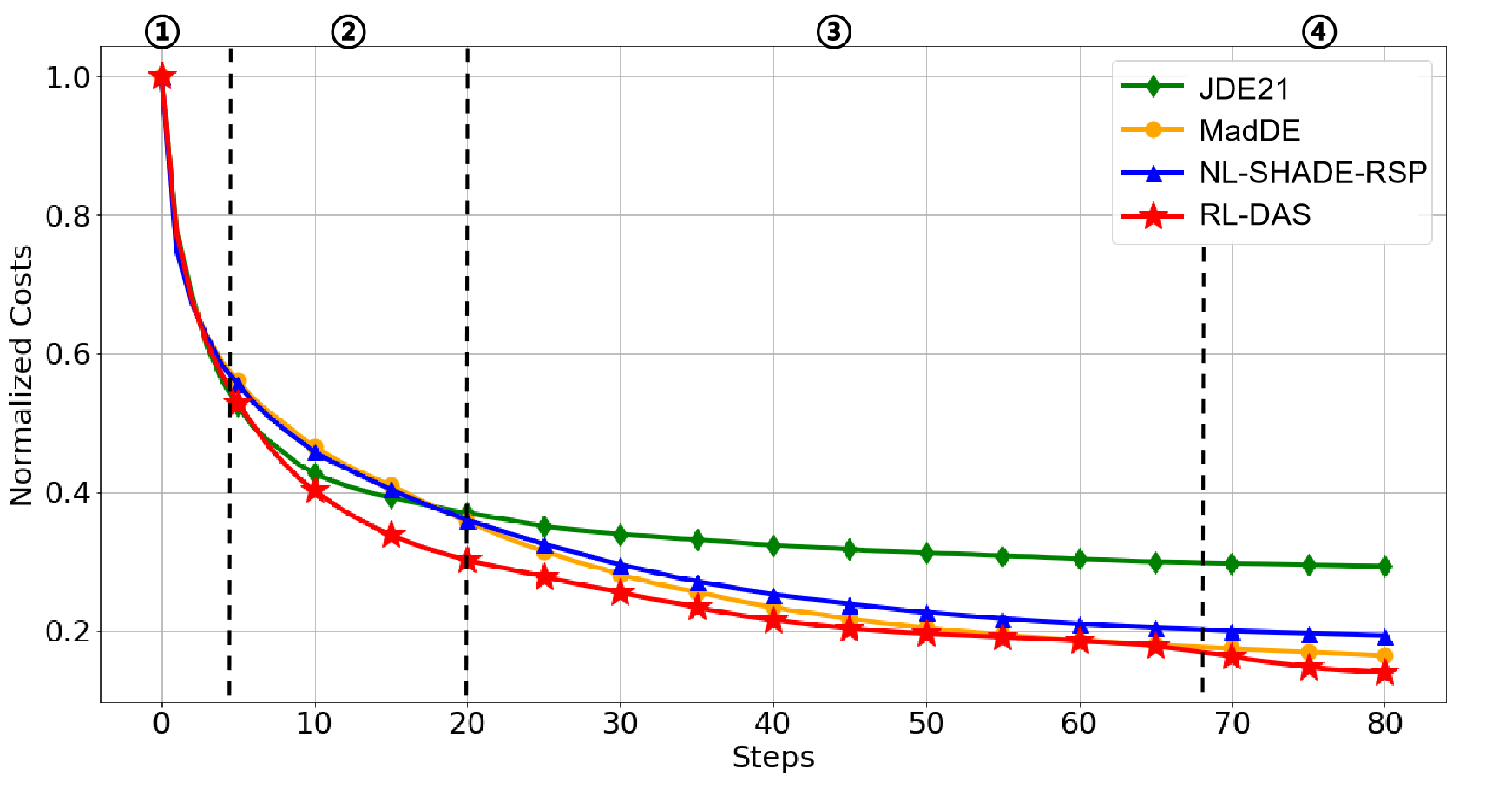}
        }
        \vspace{-2mm}
		\subfloat[]{
        \label{schwefel_dist}
		\includegraphics[width=0.4\textwidth]{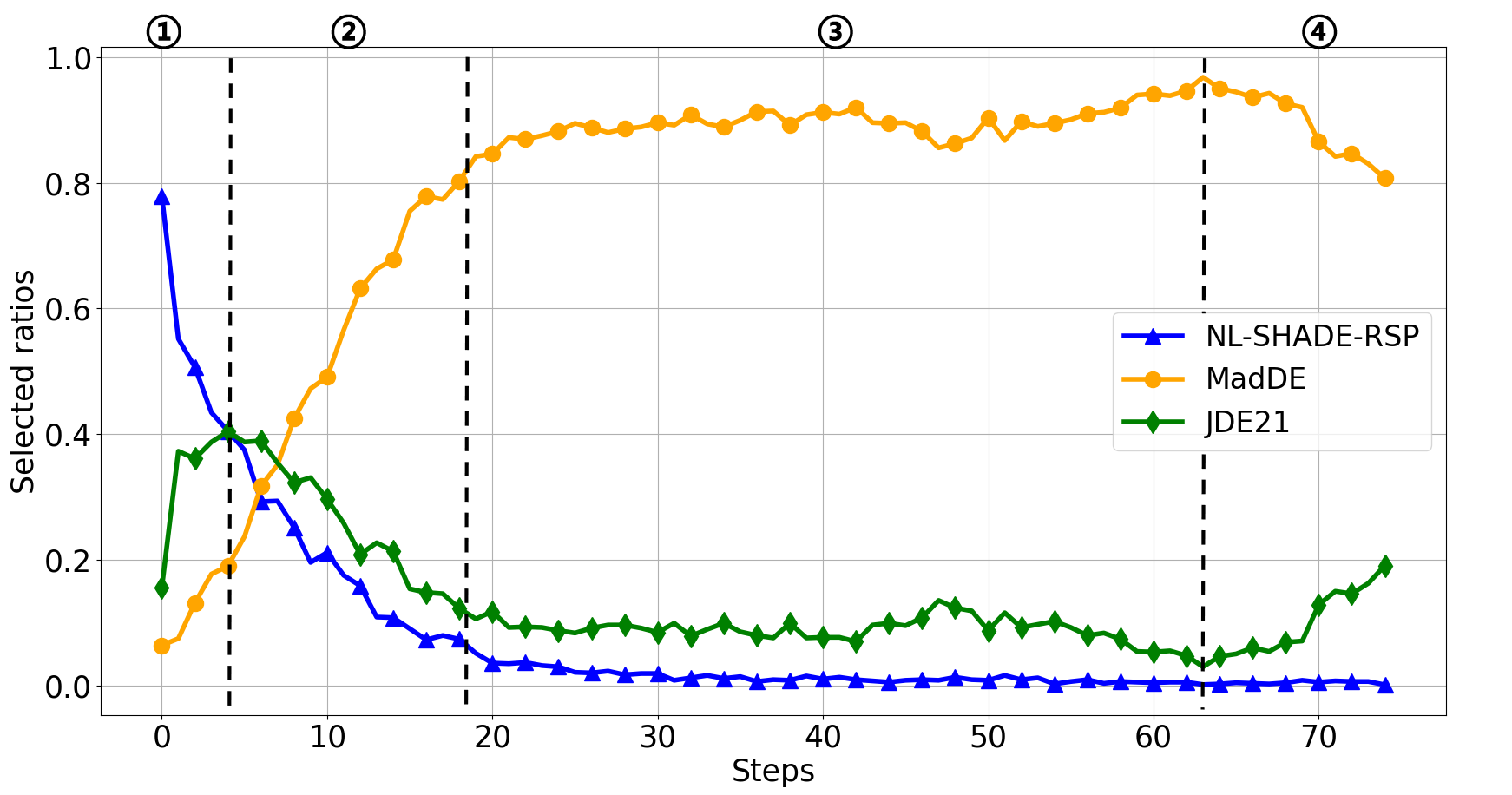}
        }
\caption{The descent curve (a) and selection distribution (b) on C2.}
\label{dist_curve}
\end{figure}

This study uses DRL to learn a policy that dynamically selects the most proper EC algorithm according to the optimization state. The experimental results in the main body of our paper have proven the efficiency of RL-DAS. This subsection conducts an in-depth analysis of the learned policy, by depicting the search behaviors of different candidate algorithms and the selecting results of RL-DAS during the optimization progress, 
where the 10D Schwefel problem is taken as an example for illustration. The results are shown in Fig.~\ref{dist_curve}. 
Specifically, Fig.~\ref{schwefel_curve} records the normalized costs obtained by different algorithms for every 2,500 function evaluations, whereas Fig.~\ref{schwefel_dist} reports the proportion of selecting each candidate algorithm in each RL step.
 Note that the abscissa of these two sub-figures cannot be completely aligned: due to additional function evaluations consumed by random walk sampling in states, the actual evaluations allocated to each RL step is larger than 2,500, therefore the total number of RL steps in Fig.~\ref{schwefel_dist} is less than the recorded steps in Fig.~\ref{schwefel_curve}.

First, from Fig.~\ref{schwefel_curve}, it can be observed that NL-SHADE-RSP searches slightly faster than the JDE21 and MadDE in the initial steps (phase 1). 
At this time, from Fig.~\ref{schwefel_dist}, the selection proportion of NL-SHADE-RSP is close to 80\%, indicating that the RL-DAS agent tries to make good use of the best performance of NL-SHADE-RSP at the beginning. But this proportion quickly decreases, 
while the selection proportion of JDE21 increases. This is because JDE21 quickly surpasses the NL-SHADE-RSP and becomes the best algorithm with its strong exploitation strategy. 
Then, as JDE21 gradually converges (phase 2), RL-DAS reduces its proportion and turns to the best algorithm, MadDE, after 20 steps (phase 3), to perform the exploratory search. Note that, at this stage, RL-DAS still retains some selections on JDE21 to achieve a balance between exploration and exploitation. This selection preference is kept 
 until the last 10 steps (phase 4), where more JDE21 options are chosen to make a sprint, indicating that the agent realizes that the optimization is about to end and hence final exploitation is required. The performance of RL-DAS in this period has a promotion that makes the final performance of RL-DAS defeat others. The experimental result proves that the agent has learned the different characteristics among candidates and is able to adapt the optimization behavior according to the environment, helping RL-DAS achieve both the fastest convergence and the highest optimization accuracy.

\begin{table}[h]
\centering
\caption{Transfer results from one class of problems to others.}
\resizebox{0.45\textwidth}{!}{%
\begin{tabular}{ccc|c|c|c}
\hline
\multicolumn{2}{c|}{\begin{tabular}[c]{@{}c@{}}Problem\\Class\end{tabular}}
& \multicolumn{1}{c|}{\begin{tabular}[c]{@{}c@{}}\textbf{RL-DAS}\\\textbf{Zero-shot (C1)}\end{tabular}}
& \multicolumn{1}{c|}{\begin{tabular}[c]{@{}c@{}}\textbf{RL-DAS}\\\textbf{Zero-shot (C2)}\end{tabular}}
& \multicolumn{1}{c|}{\begin{tabular}[c]{@{}c@{}}\textbf{RL-DAS}\\\textbf{Zero-shot (C11)}\end{tabular}}
& \multicolumn{1}{c}{\begin{tabular}[c]{@{}c@{}}\textbf{RL-DAS}\end{tabular}}

\\ \hline
\multicolumn{1}{c|}{C1}
& \multicolumn{1}{c|}{\begin{tabular}[c]{@{}c@{}}Mean\\Std\end{tabular}}
& \begin{tabular}[c]{@{}c@{}}$\backslash$\\$\backslash$\end{tabular}
& \begin{tabular}[c]{@{}c@{}}\textbf{2.058e-8}\\\textbf{2.231e-8}\end{tabular}
& \begin{tabular}[c]{@{}c@{}}6.477e-8\\3.104e-8\end{tabular}
& \begin{tabular}[c]{@{}c@{}}2.974e-8\\2.156e-8\end{tabular}

\\
\multicolumn{1}{c|}{C2}
& \multicolumn{1}{c|}{\begin{tabular}[c]{@{}c@{}}Mean\\Std\end{tabular}}
& \begin{tabular}[c]{@{}c@{}}3.726e2\\1.339e2\end{tabular}
& \begin{tabular}[c]{@{}c@{}}$\backslash$\\$\backslash$\end{tabular}
& \begin{tabular}[c]{@{}c@{}}3.876e2\\1.258e2\end{tabular}
& \begin{tabular}[c]{@{}c@{}}\textbf{3.334e2}\\\textbf{1.108e2}\end{tabular}

\\
\multicolumn{1}{c|}{C3}
& \multicolumn{1}{c|}{\begin{tabular}[c]{@{}c@{}}Mean\\Std\end{tabular}}
& \begin{tabular}[c]{@{}c@{}}1.722e1\\1.677e0\end{tabular}
& \begin{tabular}[c]{@{}c@{}}1.746e1\\1.688e0\end{tabular}
& \begin{tabular}[c]{@{}c@{}}\textbf{1.647e1}\\\textbf{1.587e0}\end{tabular}
& \begin{tabular}[c]{@{}c@{}}1.738e1\\1.666e0\end{tabular}

\\
\multicolumn{1}{c|}{C4}
& \multicolumn{1}{c|}{\begin{tabular}[c]{@{}c@{}}Mean\\Std\end{tabular}}
& \begin{tabular}[c]{@{}c@{}}1.099e0\\3.363e-1\end{tabular}
& \begin{tabular}[c]{@{}c@{}}\textbf{9.971e-1}\\\textbf{3.746e-1}\end{tabular}
& \begin{tabular}[c]{@{}c@{}}1.014e0\\4.187e-1\end{tabular}
& \begin{tabular}[c]{@{}c@{}}9.982e-1\\3.487e-1\end{tabular}

\\
\multicolumn{1}{c|}{C5}
& \multicolumn{1}{c|}{\begin{tabular}[c]{@{}c@{}}Mean\\Std\end{tabular}}
& \begin{tabular}[c]{@{}c@{}}1.623e1\\2.510e1\end{tabular}
& \begin{tabular}[c]{@{}c@{}}1.599e1\\2.228e1\end{tabular}
& \begin{tabular}[c]{@{}c@{}}1.626e1\\2.481e1\end{tabular}
& \begin{tabular}[c]{@{}c@{}}\textbf{1.587e1}\\\textbf{2.314e1}\end{tabular}

\\
\multicolumn{1}{c|}{C6}
& \multicolumn{1}{c|}{\begin{tabular}[c]{@{}c@{}}Mean\\Std\end{tabular}}
& \begin{tabular}[c]{@{}c@{}}5.844e0\\3.487e0\end{tabular}
& \begin{tabular}[c]{@{}c@{}}\textbf{4.082e0}\\\textbf{3.014e0}\end{tabular}
& \begin{tabular}[c]{@{}c@{}}5.752e0\\3.342e0\end{tabular}
& \begin{tabular}[c]{@{}c@{}}5.659e0\\3.211e0\end{tabular}

\\
\multicolumn{1}{c|}{C7}
& \multicolumn{1}{c|}{\begin{tabular}[c]{@{}c@{}}Mean\\Std\end{tabular}}
& \begin{tabular}[c]{@{}c@{}}9.854e0\\1.099e1\end{tabular}
& \begin{tabular}[c]{@{}c@{}}\textbf{9.777e0}\\\textbf{9.627e0}\end{tabular}
& \begin{tabular}[c]{@{}c@{}}9.876e0\\9.811e0\end{tabular}
& \begin{tabular}[c]{@{}c@{}}9.783e0\\1.058e1\end{tabular}

\\
\multicolumn{1}{c|}{C8}
& \multicolumn{1}{c|}{\begin{tabular}[c]{@{}c@{}}Mean\\Std\end{tabular}}
& \begin{tabular}[c]{@{}c@{}}\textbf{5.152e1}\\\textbf{1.164e1}\end{tabular}
& \begin{tabular}[c]{@{}c@{}}5.248e1\\1.394e1\end{tabular}
& \begin{tabular}[c]{@{}c@{}}5.243e1\\1.388e1\end{tabular}
& \begin{tabular}[c]{@{}c@{}}5.192e1\\1.259e1\end{tabular}

\\
\multicolumn{1}{c|}{C9}
& \multicolumn{1}{c|}{\begin{tabular}[c]{@{}c@{}}Mean\\Std\end{tabular}}
& \begin{tabular}[c]{@{}c@{}}8.053e1\\2.496e1\end{tabular}
& \begin{tabular}[c]{@{}c@{}}8.388e1\\2.798e1\end{tabular}
& \begin{tabular}[c]{@{}c@{}}8.225e1\\2.550e1\end{tabular}
& \begin{tabular}[c]{@{}c@{}}\textbf{7.988e1}\\\textbf{2.468e1}\end{tabular}

\\
\multicolumn{1}{c|}{C10}
& \multicolumn{1}{c|}{\begin{tabular}[c]{@{}c@{}}Mean\\Std\end{tabular}}
& \begin{tabular}[c]{@{}c@{}}1.787e2\\1.597e1\end{tabular} 
& \begin{tabular}[c]{@{}c@{}}\textbf{1.712e2}\\\textbf{1.581e1}\end{tabular}
& \begin{tabular}[c]{@{}c@{}}1.746e2\\1.670e1\end{tabular}
& \begin{tabular}[c]{@{}c@{}}1.715e2\\1.501e1\end{tabular}

\\ \hline
\end{tabular}%
}
\label{appx:zeroshot}
\end{table}

\begin{table}[h]
    \centering
    \caption{Transfer results under different problem class partitions.}
    \resizebox{0.37\textwidth}{!}{%
    \begin{tabular}{ccc|c|c|c}
    \hline
    \multicolumn{2}{c|}{\begin{tabular}[c]{@{}c@{}}Transfer\\Setting\end{tabular}}
    & \multicolumn{1}{c|}{\begin{tabular}[c]{@{}c@{}}\textbf{NL-SHADE}\\\textbf{-RSP}\end{tabular}}
    & \multicolumn{1}{c|}{\begin{tabular}[c]{@{}c@{}}\textbf{MadDE}\end{tabular}}
    & \multicolumn{1}{c|}{\begin{tabular}[c]{@{}c@{}}\textbf{JDE21}\end{tabular}}
    & \multicolumn{1}{c}{\begin{tabular}[c]{@{}c@{}}\textbf{RL-DAS}\\\textbf{(Zero-shot)}\end{tabular}}
    
    \\ \hline
    \multicolumn{1}{c|}{TS1}
    & \multicolumn{1}{c|}{\begin{tabular}[c]{@{}c@{}}Mean\\Std\end{tabular}}
    & \begin{tabular}[c]{@{}c@{}}7.018e1\\2.369e1\end{tabular}
    & \begin{tabular}[c]{@{}c@{}}4.457e1\\1.117e1\end{tabular}
    & \begin{tabular}[c]{@{}c@{}}6.118e1\\1.569e1\end{tabular} 
    & \begin{tabular}[c]{@{}c@{}}\textbf{4.312e1}\\\textbf{1.144e1}\end{tabular}
    
    \\
    \multicolumn{1}{c|}{TS2}
    & \multicolumn{1}{c|}{\begin{tabular}[c]{@{}c@{}}Mean\\Std\end{tabular}}
    & \begin{tabular}[c]{@{}c@{}}8.900e1\\2.960e1\end{tabular}
    & \begin{tabular}[c]{@{}c@{}}6.198e1\\1.321e1\end{tabular}
    & \begin{tabular}[c]{@{}c@{}}8.170e1\\2.589e1\end{tabular}
    & \begin{tabular}[c]{@{}c@{}}\textbf{6.110e1}\\\textbf{1.296e1}\end{tabular}
    
    \\
    \multicolumn{1}{c|}{TS3}
    & \multicolumn{1}{c|}{\begin{tabular}[c]{@{}c@{}}Mean\\Std\end{tabular}}
    & \begin{tabular}[c]{@{}c@{}}1.349e2\\2.518e1\end{tabular}
    & \begin{tabular}[c]{@{}c@{}}1.245e2\\2.347e1\end{tabular}
    & \begin{tabular}[c]{@{}c@{}}1.334e2\\2.499e1\end{tabular}
    & \begin{tabular}[c]{@{}c@{}}\textbf{1.219e2}\\\textbf{1.931e1}\end{tabular}
    
    \\
    
    
    \hline
    \end{tabular}%
    }
    \label{appx:noisy}
    \end{table}

\begin{table}[h]
\centering
\caption{The ablation study performance.}
\resizebox{0.37\textwidth}{!}{%
\begin{tabular}{cc|c|c|c|c}
\hline
\multicolumn{2}{c|}{\begin{tabular}[c]{@{}c@{}}Problem\\Class\end{tabular}}
& \multicolumn{1}{c|}{\begin{tabular}[c]{@{}c@{}}RL-DAS\\w/o LA\end{tabular}}
& \multicolumn{1}{c|}{\begin{tabular}[c]{@{}c@{}}RL-DAS\\w/o AH\end{tabular}}
& \multicolumn{1}{c|}{\begin{tabular}[c]{@{}c@{}}RL-DAS\\w/o Context\end{tabular}}
& \multicolumn{1}{c}{RL-DAS}
\\ \hline

C1
& \multicolumn{1}{|c|}{\begin{tabular}[c]{@{}c@{}}Mean\\Std\end{tabular}}
& \begin{tabular}[c]{@{}c@{}}6.872e-6\\7.361e-6\end{tabular}
& \begin{tabular}[c]{@{}c@{}}8.931e-7\\1.116e-6\end{tabular}
& \begin{tabular}[c]{@{}c@{}}4.587e6\\4.697e6\end{tabular}
& \begin{tabular}[c]{@{}c@{}}\textbf{2.974e-8}\\\textbf{2.156e-8}\end{tabular}
\\
C2
& \multicolumn{1}{|c|}{\begin{tabular}[c]{@{}c@{}}Mean\\Std\end{tabular}}
& \begin{tabular}[c]{@{}c@{}}4.208e2\\1.441e2\end{tabular}
& \begin{tabular}[c]{@{}c@{}}4.104e2\\1.534e2\end{tabular}
& \begin{tabular}[c]{@{}c@{}}1.038e3\\3.697e2\end{tabular}
& \begin{tabular}[c]{@{}c@{}}\textbf{3.334e2}\\\textbf{1.108e2}\end{tabular}
\\
C3
& \multicolumn{1}{|c|}{\begin{tabular}[c]{@{}c@{}}Mean\\Std\end{tabular}}
& \begin{tabular}[c]{@{}c@{}}1.865e1\\1.838e0\end{tabular}
& \begin{tabular}[c]{@{}c@{}}1.845e1\\1.748e0\end{tabular}
& \begin{tabular}[c]{@{}c@{}}3.909e1\\5.201e0\end{tabular}
& \begin{tabular}[c]{@{}c@{}}\textbf{1.738e1}\\\textbf{1.666e0}\end{tabular}
\\
C4
& \multicolumn{1}{|c|}{\begin{tabular}[c]{@{}c@{}}Mean\\Std\end{tabular}}
& \begin{tabular}[c]{@{}c@{}}1.157e0\\5.069e-1\end{tabular}
& \begin{tabular}[c]{@{}c@{}}1.143e0\\4.987e-1\end{tabular}
& \begin{tabular}[c]{@{}c@{}}2.698e2\\9.697e1\end{tabular}
& \begin{tabular}[c]{@{}c@{}}\textbf{9.982e-1}\\\textbf{3.487e-1}\end{tabular}
\\
C5
& \multicolumn{1}{|c|}{\begin{tabular}[c]{@{}c@{}}Mean\\Std\end{tabular}}
& \begin{tabular}[c]{@{}c@{}}1.603e1\\2.669e1\end{tabular}
& \begin{tabular}[c]{@{}c@{}}1.618e1\\2.589e1\end{tabular}
& \begin{tabular}[c]{@{}c@{}}6.972e2\\9.676e2\end{tabular}
& \begin{tabular}[c]{@{}c@{}}\textbf{1.587e1}\\\textbf{2.314e1}\end{tabular}
\\
C6
& \multicolumn{1}{|c|}{\begin{tabular}[c]{@{}c@{}}Mean\\Std\end{tabular}}
& \begin{tabular}[c]{@{}c@{}}9.648e0\\6.033e0\end{tabular}
& \begin{tabular}[c]{@{}c@{}}1.011e1\\6.352e0\end{tabular}
& \begin{tabular}[c]{@{}c@{}}4.564e1\\2.112e1\end{tabular}
& \begin{tabular}[c]{@{}c@{}}\textbf{5.659e0}\\\textbf{3.211e0}\end{tabular}
\\
C7
& \multicolumn{1}{|c|}{\begin{tabular}[c]{@{}c@{}}Mean\\Std\end{tabular}}
& \begin{tabular}[c]{@{}c@{}}3.145e1\\3.672e1\end{tabular}
& \begin{tabular}[c]{@{}c@{}}1.593e1\\1.649e1\end{tabular}
& \begin{tabular}[c]{@{}c@{}}2.331e2\\2.522e2\end{tabular}
& \begin{tabular}[c]{@{}c@{}}\textbf{9.783e0}\\\textbf{1.058e1}\end{tabular}
\\
C8
& \multicolumn{1}{|c|}{\begin{tabular}[c]{@{}c@{}}Mean\\Std\end{tabular}}
& \begin{tabular}[c]{@{}c@{}}5.500e1\\1.415e1\end{tabular}
& \begin{tabular}[c]{@{}c@{}}5.381e1\\1.399e1\end{tabular}
& \begin{tabular}[c]{@{}c@{}}8.397e1\\2.690e1\end{tabular}
& \begin{tabular}[c]{@{}c@{}}\textbf{5.192e1}\\\textbf{1.259e1}\end{tabular}
\\
C9
& \multicolumn{1}{|c|}{\begin{tabular}[c]{@{}c@{}}Mean\\Std\end{tabular}}
& \begin{tabular}[c]{@{}c@{}}8.560e1\\2.971e1\end{tabular}
& \begin{tabular}[c]{@{}c@{}}8.282e1\\2.566e1\end{tabular}
& \begin{tabular}[c]{@{}c@{}}1.611e2\\5.998e1\end{tabular}
& \begin{tabular}[c]{@{}c@{}}\textbf{7.988e1}\\\textbf{2.468e1}\end{tabular}
\\
C10
& \multicolumn{1}{|c|}{\begin{tabular}[c]{@{}c@{}}Mean\\Std\end{tabular}}
& \begin{tabular}[c]{@{}c@{}}1.799e2\\1.680e1\end{tabular}
& \begin{tabular}[c]{@{}c@{}}1.873e2\\1.659e1\end{tabular}
& \begin{tabular}[c]{@{}c@{}}3.416e2\\4.644e1\end{tabular}
& \begin{tabular}[c]{@{}c@{}}\textbf{1.715e2}\\\textbf{1.501e1}\end{tabular}
\\
C11
& \multicolumn{1}{|c|}{\begin{tabular}[c]{@{}c@{}}Mean\\Std\end{tabular}}
& \begin{tabular}[c]{@{}c@{}}8.592e1\\3.786e1\end{tabular}
& \begin{tabular}[c]{@{}c@{}}8.337e1\\3.668e1\end{tabular}
& \begin{tabular}[c]{@{}c@{}}1.917e2\\7.250e1\end{tabular}
& \begin{tabular}[c]{@{}c@{}}\textbf{7.386e1}\\\textbf{2.364e1}\end{tabular}
\\ \hline
\end{tabular}%
}
\label{appx:ablation}
\end{table}

\begin{table}[h]
\centering
\caption{The optimization performance under different reward schemes.}
\resizebox{0.39\textwidth}{!}{%
\begin{tabular}{cc|ccccc}
\hline
 \multicolumn{2}{c|}{\multirow{2}{*}{\begin{tabular}[c]{@{}c@{}}Problem\\Class\end{tabular}}} 
 & \multicolumn{5}{c}{Reward Type}
\\ \cline{3-7}
&
& r1
& r2
& r3
& r4
& Ours
\\ \hline
\multirow{1}{*}{C1}
& Mean
&6.894e-8   
&\textbf{2.182e-8}   
&7.565e-8
&7.553e-8
&2.974e-8
\\
& Std
& 6.289e-8
& \textbf{1.837e-8}
& 7.462e-8
& 7.361e-8
& 2.156e-8
\\ \hline
\multirow{2}{*}{C2}
& Mean
& 3.377e2
& 3.391e2
& 3.455e2
& 3.812e2
& \textbf{3.334e2}
\\
& Std
& 1.298e2
& 1.295e2
& 1.259e2
& 1.441e2
& \textbf{1.108e2}
\\ \hline
\multirow{2}{*}{C3}
& Mean
& 1.845e1
& 1.841e1
& 1.857e1
& 2.331e1
& \textbf{1.738e1}
\\
& Std
& 1.737e0
& 1.716e0
& 1.677e0
& 1.898e0
& \textbf{1.666e0}
\\ \hline
\multirow{2}{*}{C4}
& Mean
& 1.116e0
& 1.088e0
& 1.171e0
& 1.128e0
& \textbf{9.982e-1}
\\
& Std
& 4.195e-1
& 3.866e-1
& 4.359e-1
& 4.514e-1
& \textbf{3.487e-1}
\\ \hline
\multirow{2}{*}{C5}
& Mean
& 1.777e1
& 1.690e1
& 1.797e1
& 1.861e1
& \textbf{1.587e1}
\\
& Std
& 2.311e1
& 2.396e1
& 2.557e1
& 2.564e1
& \textbf{2.314e1}
\\ \hline
\multirow{2}{*}{C6}
& Mean
& 9.184e0
& 1.429e1
& 1.225e1
& 1.939e1
& \textbf{5.659e0}
\\
& Std
& 5.414e0
& 8.333e0
& 6.122e0
& 9.935e0
& \textbf{3.211e0}
\\ \hline
\multirow{2}{*}{C7}
& Mean
& 1.132e1
& 9.852e0
& 1.006e1
& 1.102e1
& \textbf{9.783e0}
\\
& Std
& 1.224e1
& 1.069e1
& 1.103e1
& 1.206e1
& \textbf{1.058e1}
\\ \hline
\multirow{2}{*}{C8}
& Mean
& 5.224e1
& 5.317e1
& 5.381e1
& 5.205e1
& \textbf{5.192e1}
\\
& Std
& 1.219e1
& 1.314e1
& 1.355e1
& 1.268e1
& \textbf{1.259e1}
\\ \hline
\multirow{2}{*}{C9}
& Mean
& 8.004e1
& 8.282e1
& 8.388e1
& 8.846e1
& \textbf{7.988e1}
\\
& Std
& 2.317e1
& 2.647e1
& 2.591e1
& 2.970e1
& \textbf{2.468e1}
\\ \hline
\multirow{2}{*}{C10}
& Mean
& 1.742e2
& 1.752e2
& 1.811e2
& 1.892e2
& \textbf{1.715e2}
\\
& Std
& 1.584e1
& 1.543e1
& 1.682e1
& 1.675e1
& \textbf{1.501e1}
\\ \hline
\multirow{2}{*}{C11}
& Mean
& 7.740e1
& 7.849e1
& 7.898e1
& 8.446e1
& \textbf{7.386e1}
\\
& Std
& 2.410e1
& 2.938e1
& 2.863e1
& 3.217e1
& \textbf{2.364e1}
\\ \hline
\end{tabular}%
}
\label{appx:reward}
\end{table}

\section{Absolute Cost Values}
In the main body of our paper, we report the cost decent in percentage of different algorithms from the random initialization. This section additionally presents the detailed absolute results of the algorithms, where the mean and standard deviations are presented. Specifically, Table~\ref{appx:r+s} and Table~\ref{appx:20d} supplement the results for major performance comparisons of RL-DAS and the competitors. Table~\ref{appx:zeroshot} and Table~\ref{appx:noisy} shows the detailed results of the zero-shot transfer experiment. Table~\ref{appx:ablation} is the results for the ablation study on framework components, while Table~\ref{appx:reward} delves into the results of the reward study.
\vfill
\newpage
\bibliographystyle{IEEEtran}
\bibliography{References}

\end{document}